\documentclass[fleqn,10pt]{wlscirep}
\usepackage[utf8]{inputenc}
\usepackage[T1]{fontenc}
\usepackage{graphicx}
\usepackage{xcolor}
\usepackage{hyperref}
\usepackage{subcaption}

\title{Knowledge Graph Extraction from Biomedical Literature for Alkaptonuria Rare Disease}

\author[1,*]{Giang Pham}
\author[2]{Rebecca Finetti}
\author[3]{Caterina Graziani}
\author[2]{Bianca Roncaglia}
\author[3]{Asma Bendjeddou}
\author[4]{Linda Brodo}
\author[3]{Sara Brunetti}
\author[3]{Moreno Falaschi}
\author[1] {Stefano Forti}
\author[1]{Silvia Giulia Galfré}
\author[1]{Paolo Milazzo}
\author[1]{Corrado Priami}
\author[2]{Annalisa Santucci}
\author[2]{Ottavia Spiga}
\author[5]{Alina  S\^irbu}
\affil[1]{Department of Computer Science, University of Pisa, Pisa, Italy.}
\affil[2]{Department of Biotechnology, Chemistry and Pharmacy, University of Siena, Siena, Italy}
\affil[3]{Department of Information Engineering and Mathematics, University of Siena, Siena, Italy}
\affil[4]{Department of Scienze Economiche e Aziendali, University of Sassari, Sassari, Italy.}
\affil[5]{Department of Computer Science and Engineering, University of Bologna, Bologna, Italy.}

\affil[*]{tranhuong.giangpham@phd.unipi.it}

\begin{abstract}
Alkaptonuria (AKU) is an ultra-rare autosomal recessive metabolic disorder caused by mutations in the HGD (Homogentisate 1,2-Dioxygenase) gene, leading to a pathological accumulation of homogentisic acid
(HGA) in body fluids and tissues. This leads to systemic manifestations, including premature spondyloarthropathy, renal and prostatic stones, and cardiovascular complications. Being ultra-rare, the amount of data related to the disease is limited, both in terms of clinical data and literature. Knowledge graphs (KGs) can help connect the limited knowledge about the disease (basic mechanisms, manifestations and existing therapies) with other knowledge; however, AKU is frequently underrepresented or entirely absent in existing biomedical KGs. In this work, we apply a text-mining methodology based on PubTator3 for large-scale extraction of biomedical relations. 
We construct two KGs of different sizes, validate them using existing biochemical knowledge and use them to extract genes, diseases and therapies possibly related to AKU.  This computational framework reveals the systemic interactions of the disease, its comorbidities, and potential therapeutic targets, demonstrating the efficacy of our approach in analyzing rare metabolic disorders. 
\end{abstract}
\begin{document}

\flushbottom
\maketitle
%
%
\thispagestyle{empty}

\section*{Introduction}

Alkaptonuria (AKU) is an ultra-rare autosomal recessive metabolic disorder. This disease is caused by mutations in the HGD gene, which encodes the enzyme homogentisate 1,2-dioxygenase. This enzyme is involved in the breakdown of tyrosine and phenylalanine \cite{fernandez1996molecular}. Loss or severe reduction of HGD activity interrupts this metabolic pathway, causing a pathological accumulation of homogentisic acid (HGA) in body fluids and tissues. Once accumulated, HGA undergoes oxidation and radical polymerization into a melanin-like polymer \cite{chow2020pigmentation}. This process, known as ochronosis, is characterized by pigment deposits in collagen-rich tissues such as cartilage, heart valves, tendons, and ligaments. Pigment deposition is accompanied by chronic oxidative stress, persistent inflammation, and secondary amyloidosis \cite{millucci2012alkaptonuria}, which collectively contribute to systemic manifestations including premature spondyloarthropathy, renal and prostatic stones, and cardiovascular complications \cite{bernardini2024alkaptonuria}.  

Being an ultra-rare disease, biomedical data on AKU is limited. Some clinical data exists, together with information on mutations and severity of disease in relatively reduced cohorts of patients~\cite{braconi2022effects,spiga2021machine,galderisi2022homogentisic}. Linking existing knowledge with other biomedical information can be a means to expand the understanding of the disease and produce new hypotheses in terms of therapy targets and possible new or re-purposed drugs.  Biomedical text mining has already proven effective in multiple research domains \cite{polajnar2006survey}, including drug discovery, digital epidemiology, systematic review automation \cite{weissenbacher2023text}, and visualization of large biomedical corpora \cite{costa2021exploring}.
Recently, Tsuru et al \cite{tsutsui2025literature} introduced a context-aware literature-mining framework to construct gene regulatory networks integrating BioBERT-based semantic similarity with relation extraction tools such as PubTator3\cite{wei2024pubtator} and BioREX\cite{lai2023biorex}. Kafkas et al. \cite{kafkas2019ontology} introduced one of the first ontology-guided systems for extracting pathogen–disease relationships from PubMed abstracts, producing a structured resource of over 3,000 associations to support infectious disease research. These studies collectively show how literature mining can facilitate the identification of genes, proteins, diseases, and chemical entities, creating knowledge graphs (KGs) and offering a scalable and cost-efficient alternative to manual data curation.

AKU is frequently underrepresented or entirely absent in existing biomedical KGs. Most publicly available KGs and benchmarks focus on common diseases with abundant literature and curated data, while neglecting rare conditions that lack large-scale datasets \cite{zhou2024tarkg,himmelstein2017systematic,himmelstein2015heterogeneous,whirl2021evidence}. This omission limits the ability of current KG-based models to capture the molecular mechanisms and therapeutic opportunities related to AKU.

To address these gaps in rare diseases like AKU, we developed a text-mining methodology based on PubTator3 for large-scale extraction of biomedical relations. 


The methodology includes KG \emph{creation}, \emph{validation} and \emph{knowledge extraction}.

The KG \emph{creation} step starts with PubMed search seeds related to the rare disease under analysis, extraction of entities and creation of an initial interaction network. The process is then repeated for entities closest to the initial seed. This results in a large \emph{extended network} of interactions. A smaller \emph{high-confidence network} is then extracted by keeping only edges supported by at least two articles, and removing disconnected nodes. In contrast to approaches limited to gene–gene regulation, our work includes associations among genes, diseases, and chemicals (drugs and metabolites).

KG \emph{validation} compares the two KGs against existing knowledge from biomedical databases, including STRING\cite{Szklarczyk2024STRING}, the Drug–Gene Interaction Database (DGIdb)~\cite{cannon2024dgidb}, and KEGG\cite{kanehisa2000kegg}.


\emph{Knowledge extraction}, instead, uses graph theory to extract information on important entities related to the rare disease under analysis, including an analysis of new interactions not present in existing databases.

We apply the methodology on AKU and provide a thorough interpretation of results, with the aim to extend the understanding of the disease. Our KG describes AKU as driven by a core metabolic mechanism, well known in the literature, but which connects to systemic interactions and comorbidities. Important pathways that appear to be connected are oxidative stress and inflammatory pathways, with genes such as TTR and TNF involved. Bone remodeling also emerges as important. All these could provide new therapeutic avenues to control AKU symptoms. Possible side effects of therapy and connection to neurological disorders are also suggested. All in all, our results demonstrate the efficacy of such approaches in analyzing rare metabolic disorders and generating new hypotheses for future work. 

\section*{Results}
\begin{figure}
\includegraphics[width=\textwidth]{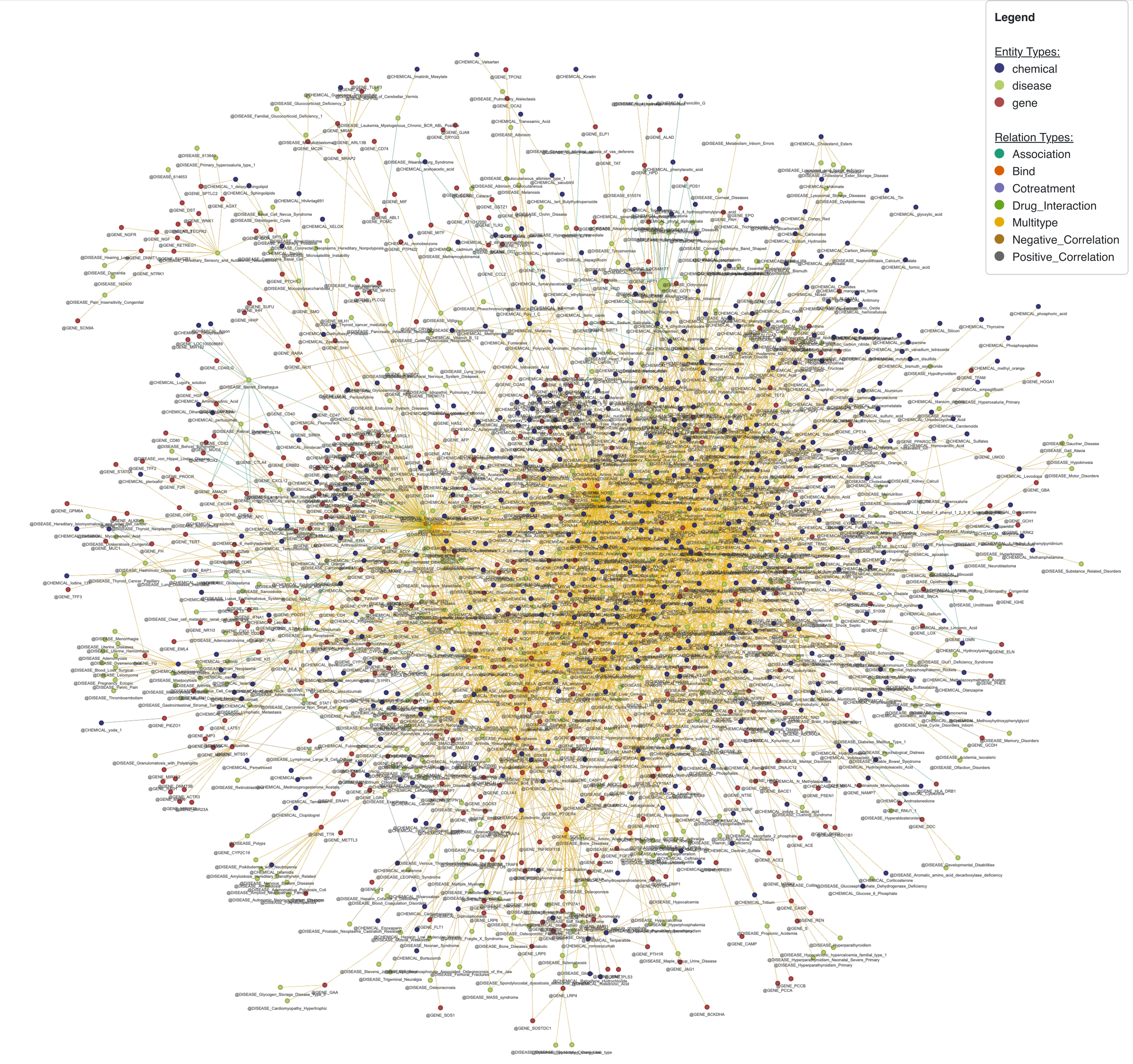}
\caption{ The \textit{high-confidence network}: KG containing only relations that appear in at least two publications. An HTML view of the network is available at the  \href{https://giangpth.github.io/Alkaptonuria/visualizations/highconfidence.html}{link}. }
\label{fig:highconf}
\end{figure}

Two KGs were extracted from PubMed, using PubTator3 for annotations. These are two undirected, weighted graphs: the \textit{extended network} and the \textit{high-confidence network} (see Methods for details). Nodes in these networks belong to the categories \textit{gene}, \textit{disease}, \textit{chemical}, and \textit{gene variant}, while edges represent relationships of types including \textit{positive correlation}, \textit{negative correlation}, \textit{association}, \textit{co-treatment}, \textit{bind}, \textit{drug interaction}, or \textit{multitype} (when two entities are connected by more than one type of relationship). The \textit{extended network} contains 27,252 nodes and 261,649 edges. The \textit{high-confidence network} was derived from the \textit{extended network} by retaining only connections supported by at least two distinct research publications, and removing isolated nodes. It consists of 1,450 nodes and 3,307 edges, and is shown in Figure \ref{fig:highconf}.  Both networks are available in JSON format as supplementary data. In the following, we validate them and use them to
extract genes, diseases and therapies possibly related to AKU.

\subsection*{KG validation}

To validate the two KGs, we compared them with graphs extracted from existing public databases: gene-gene interactions from  STRING , drug-gene interactions from DGIdb, pathways from KEGG.  For the extended network, which is quite large, we only consider the network module containing AKU in the comparison (see Methods for details).

\begin{figure}[h!] 
  \centering
  \begin{subfigure}{.48\linewidth}
    \centering
    \includegraphics[width=.98\linewidth]{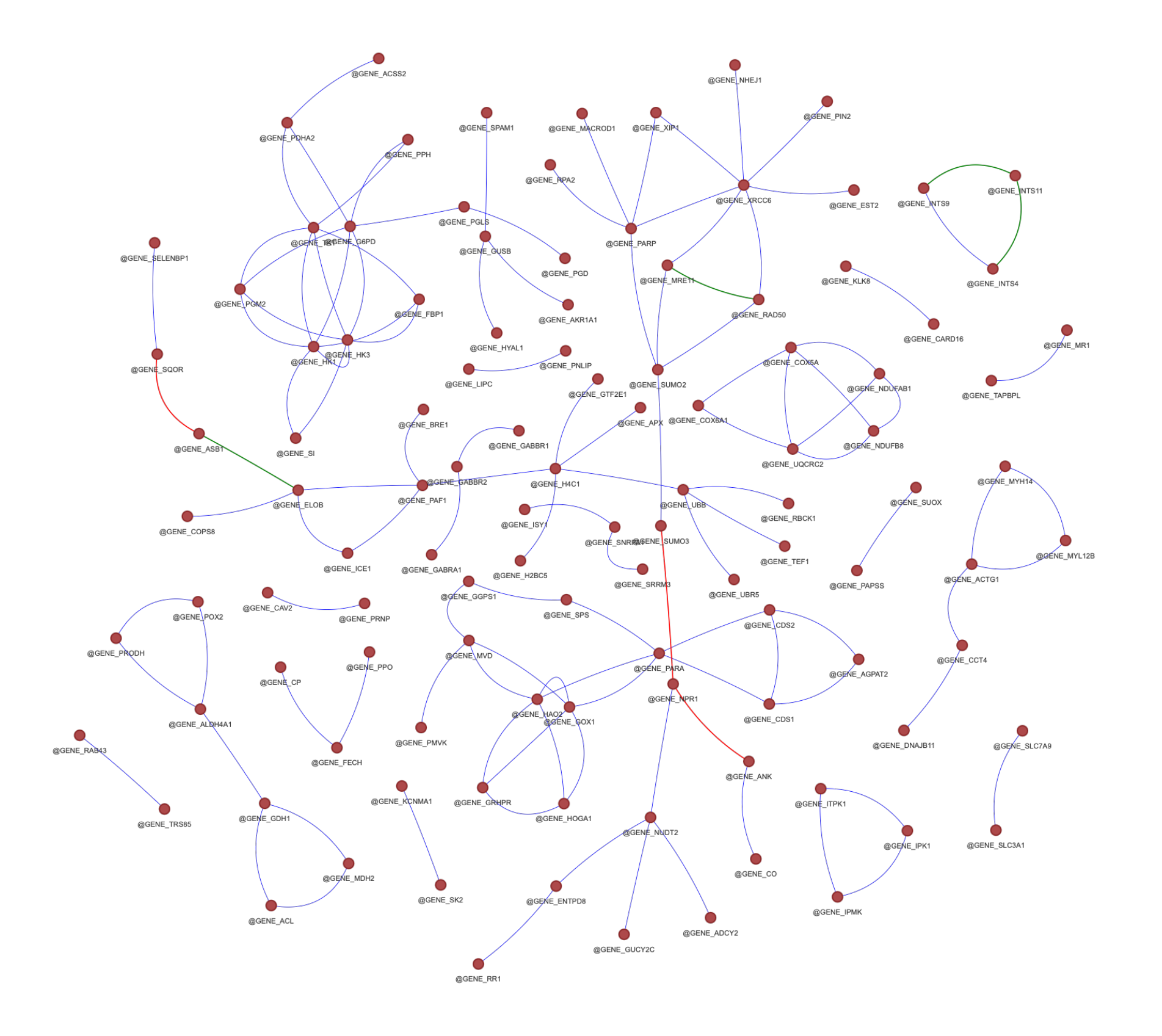}
    \caption{Visualization of gene–gene connections derived from \textit{STRING} (green and blue) and from the module containing \textit{Alkaptonuria}, \textit{HGD} and \textit{Homogentisic acid} in the \textit{extended network} (green and red). An HTML version is available at \href{https://giangpth.github.io/Alkaptonuria/visualizations/exstringdb.html}{link}.}
    \label{fig:exstringnet}
  \end{subfigure}
  \vspace{0.6em} 
  \begin{subfigure}{.48\linewidth}
    \centering
    \includegraphics[width=.98\linewidth]{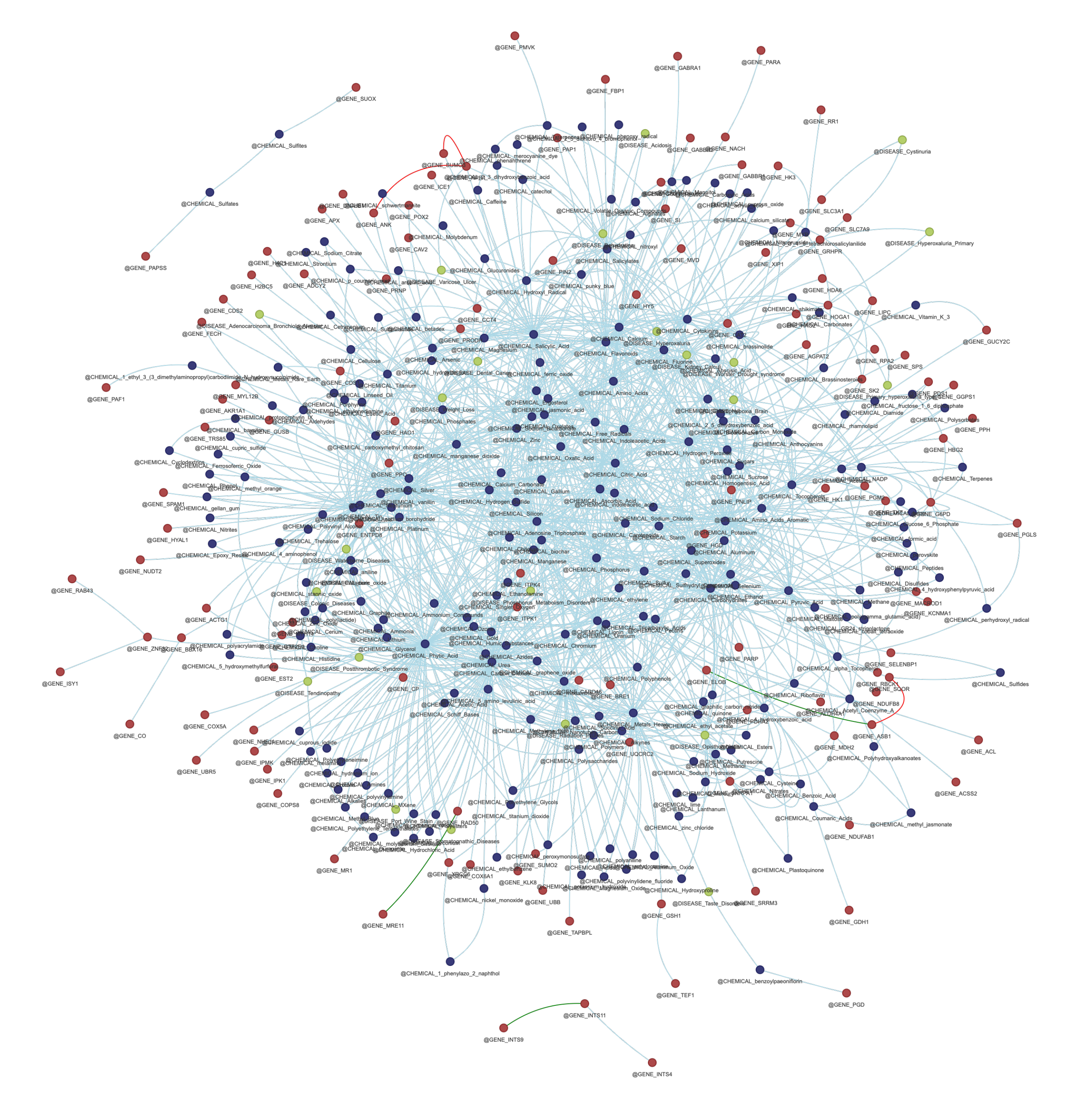}
    \caption{
    Subgraph constructed from the shortest paths in the AKU module of the \textit{extended network}, corresponding to the edges present in the \textit{ STRING } graph (Figure~\ref{fig:exstringnet}). When a direct connection (e.g., A–B) exists in \textit{ STRING } but is absent in the \textit{extended network}, the shortest indirect paths between A and B through intermediate nodes are visualized here. An HTML version is available at \href{https://giangpth.github.io/Alkaptonuria/visualizations/exstringpaths.html}{link}.}
    \label{fig:exstringpaths}
  \end{subfigure}
\begin{subfigure}{.48\linewidth}
    \centering
    \includegraphics[width=.98\linewidth]{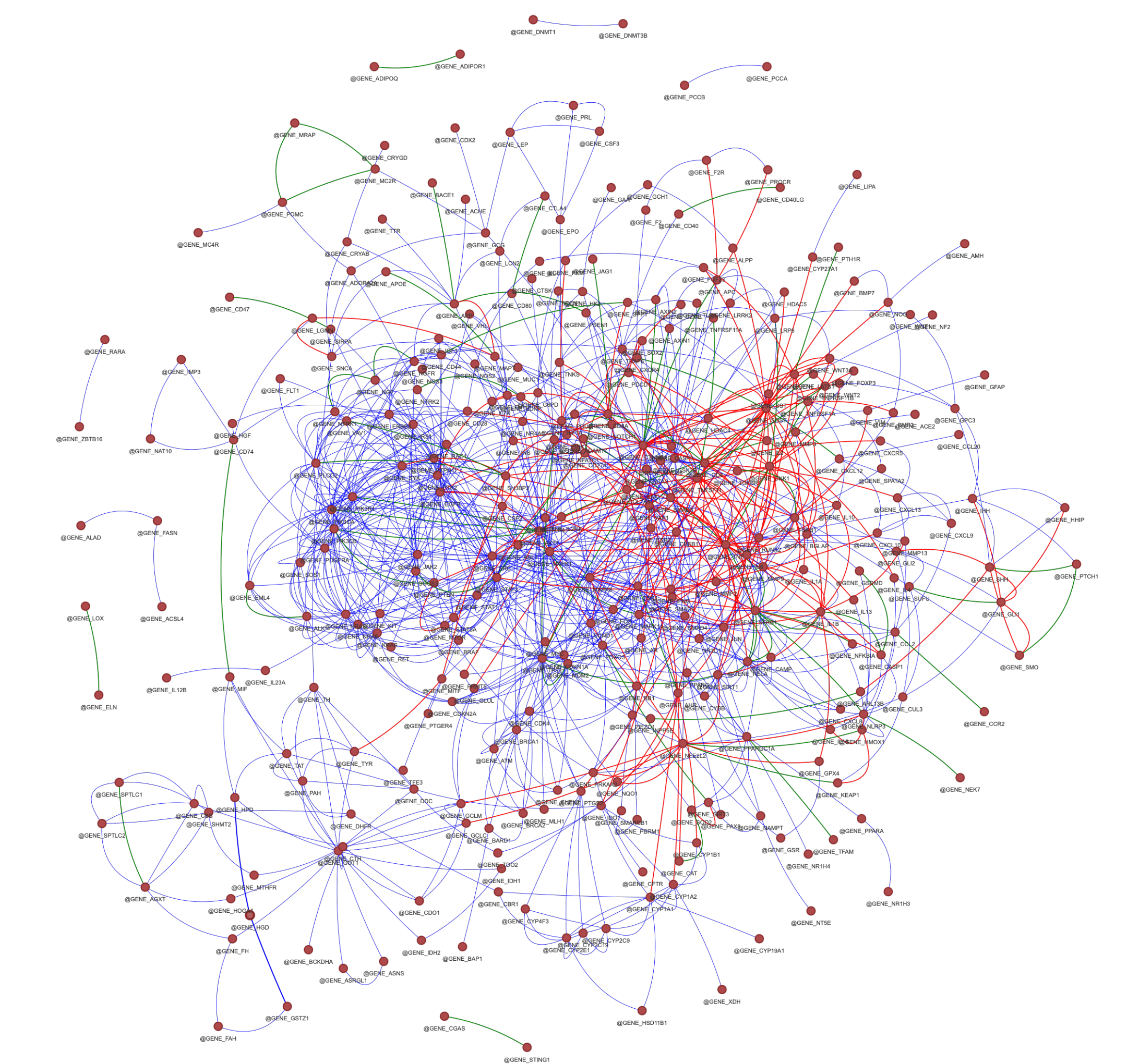}
        \caption{Visualization of gene–gene connections obtained from \textit{ STRING } (green and blue) and from the \textit{high-confidence network} (green and red). An HTML version is available at \href{https://giangpth.github.io/Alkaptonuria/visualizations/stringhi.html}{link}.}
        \label{fig:histring}
    \end{subfigure}
    \hfill
    \begin{subfigure}{.48\linewidth}
    \centering
    \includegraphics[width=.98\linewidth]{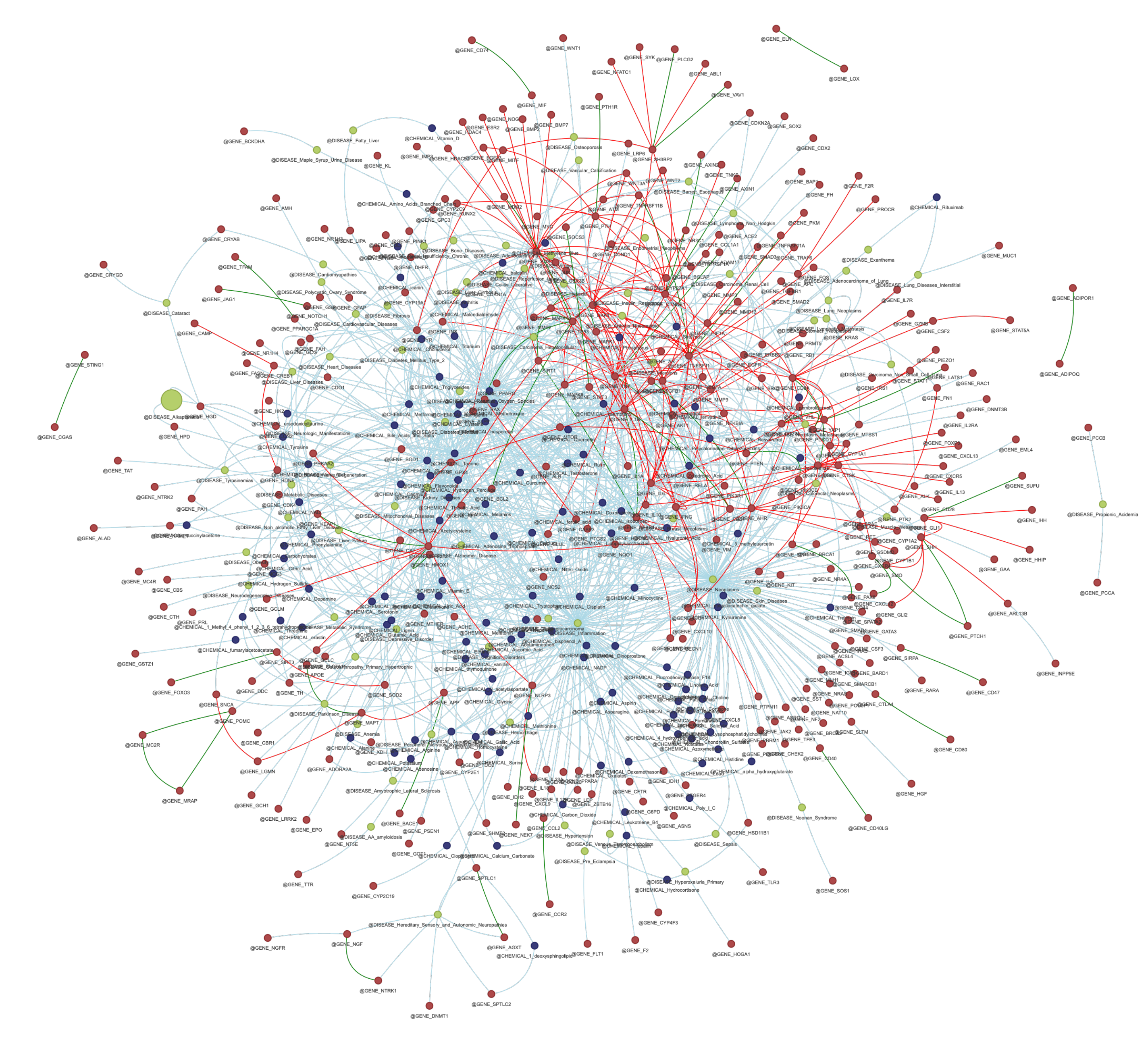}
        \caption{Subgraph constructed from the shortest paths in the \textit{high-confidence network} corresponding to the edges from the \textit{ STRING } graph shown in Figure~\ref{fig:histring}. An HTML version is available at \href{https://giangpth.github.io/Alkaptonuria/visualizations/stringhipaths.html}{link}.}
        \label{fig:histringpaths}
    \end{subfigure}

  \caption{Comparison between the two KGs  and the gene–gene connections from \textit{ STRING }.}
  \label{fig:string}
\end{figure}

Figure~\ref{fig:exstringnet} provides a comparative visualization of the module from the extended network that contains AKU, where the nodes represent genes shared between the  STRING  database and the module. Edges common to both networks are shown in green, those unique to STRING in dark blue, and those exclusive to the \textit{extended network} in red. Within this module, 3 red edges (2.5\%), 4 green edges (3.3\%), and 115 blue edges (94.2\%) were identified. The low coverage of our network of direct gene-gene interactions from  STRING  is due to the structure of our KGs: rarely genes are directly connected, but they usually interact through intermediate nodes (chemicals or diseases). To account for this, we extracted the shortest path for all pairs of directly connected genes in  STRING . Figure~\ref{fig:exstringpaths} highlights, in light blue,  the shortest paths in the module corresponding to the dark blue edges in Figure~\ref{fig:exstringnet}. Edges without corresponding paths remain dark blue. Notably, all dark blue connections are linked through intermediate nodes, resulting in no dark blue edges, indicating that our KG covers completely the knowledge in  STRING , however, with indirect interactions. The average shortest path length associated with these dark blue edges is 4.1, so direct gene-gene interaction in  STRING  is, on average, covered by a path with 3 intermediary nodes in our extended network. The low direct overlap with  STRING  suggests that the pathophysiology of AKU is highly metabolic, with the majority of functional connections between genes mediated by metabolic flux rather than a direct interaction. Nevertheless, the fact that all edges in  STRING  are recovered in its corresponding indirect shortest paths suggests the model accurately captures the multi-step enzymatic cascades present in the tyrosine pathway. This indicates that HGD deficiency has global effects, with HGA acting systemically rather than via local pathways in the interactome.

The same comparison was carried out for the \textit{high-confidence network}, as shown in Figures~\ref{fig:histring} and \ref{fig:histringpaths}. In this case, 187 red edges (17.5\%), 71 green edges (6.6\%), and 810 dark blue edges (75.9\%) were identified. Again, most blue edges could be replaced by shortest paths, with the average shortest path length at 3.82. Despite stricter filtering, the \textit{high-confidence network} retains most  STRING  interactions via short, indirect paths. This confirms that biologically meaningful gene relationships remain accessible, even under conservative evidence constraints. The reduced path length compared to the \textit{extended network} reflects a denser, more specific functional neighborhood around AKU-related genes.

\begin{figure}[h!]
    \centering
   \begin{subfigure}{.48\linewidth}
    \centering
    \includegraphics[width=.98\linewidth]{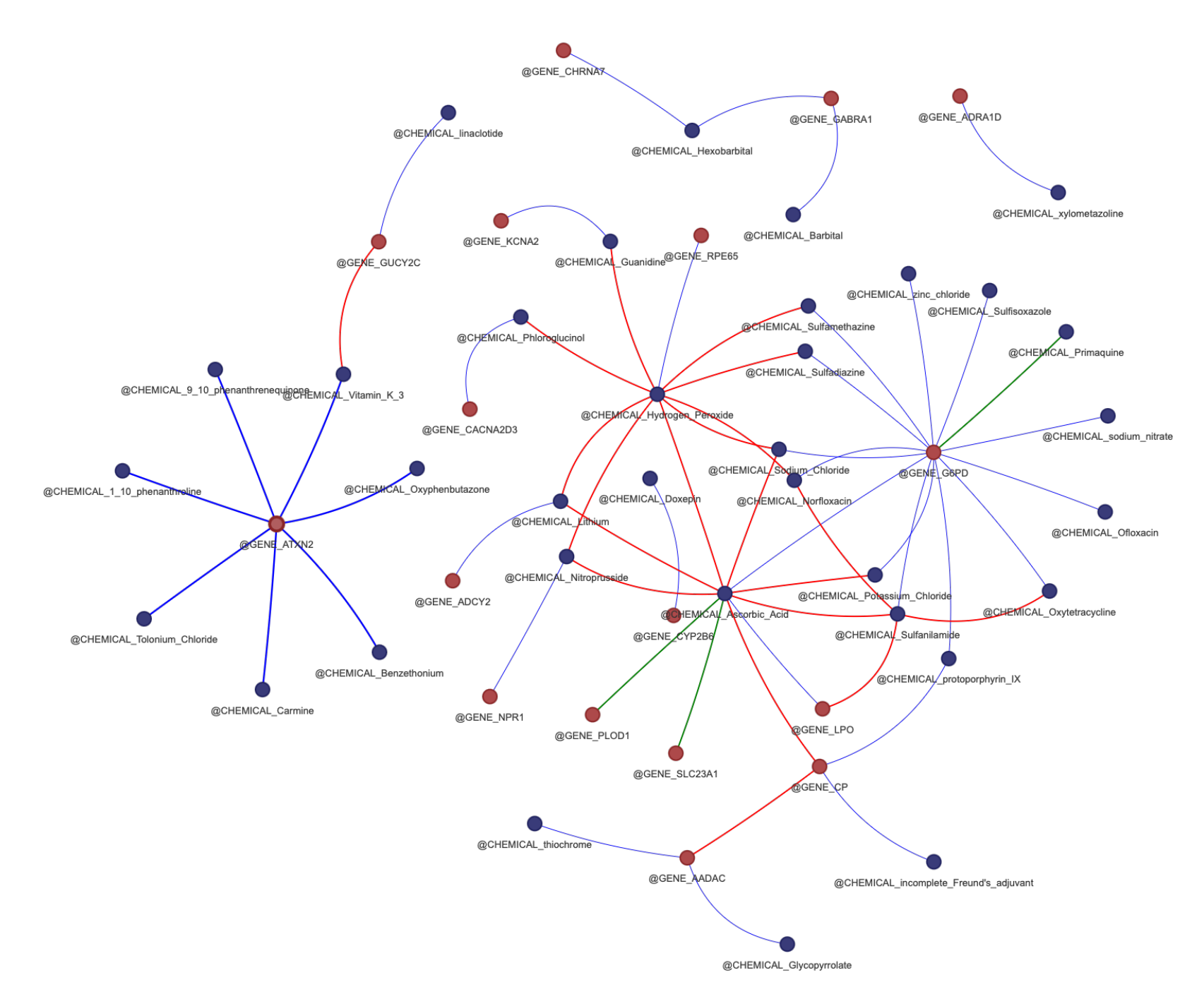}
        \caption{Visualization of gene–drug connections obtained from \textit{DGIdb} (green and blue) and from the module containing \textit{Alkaptonuria} in the \textit{extended network} (green and red). An HTML version is available at \href{https://giangpth.github.io/Alkaptonuria/visualizations/exdgidb.html}{link }.}
        \label{fig:exdgidbgraph}
    \end{subfigure}
    \hfill
    \begin{subfigure}{.48\linewidth}
    \centering
    \includegraphics[width=.98\linewidth]{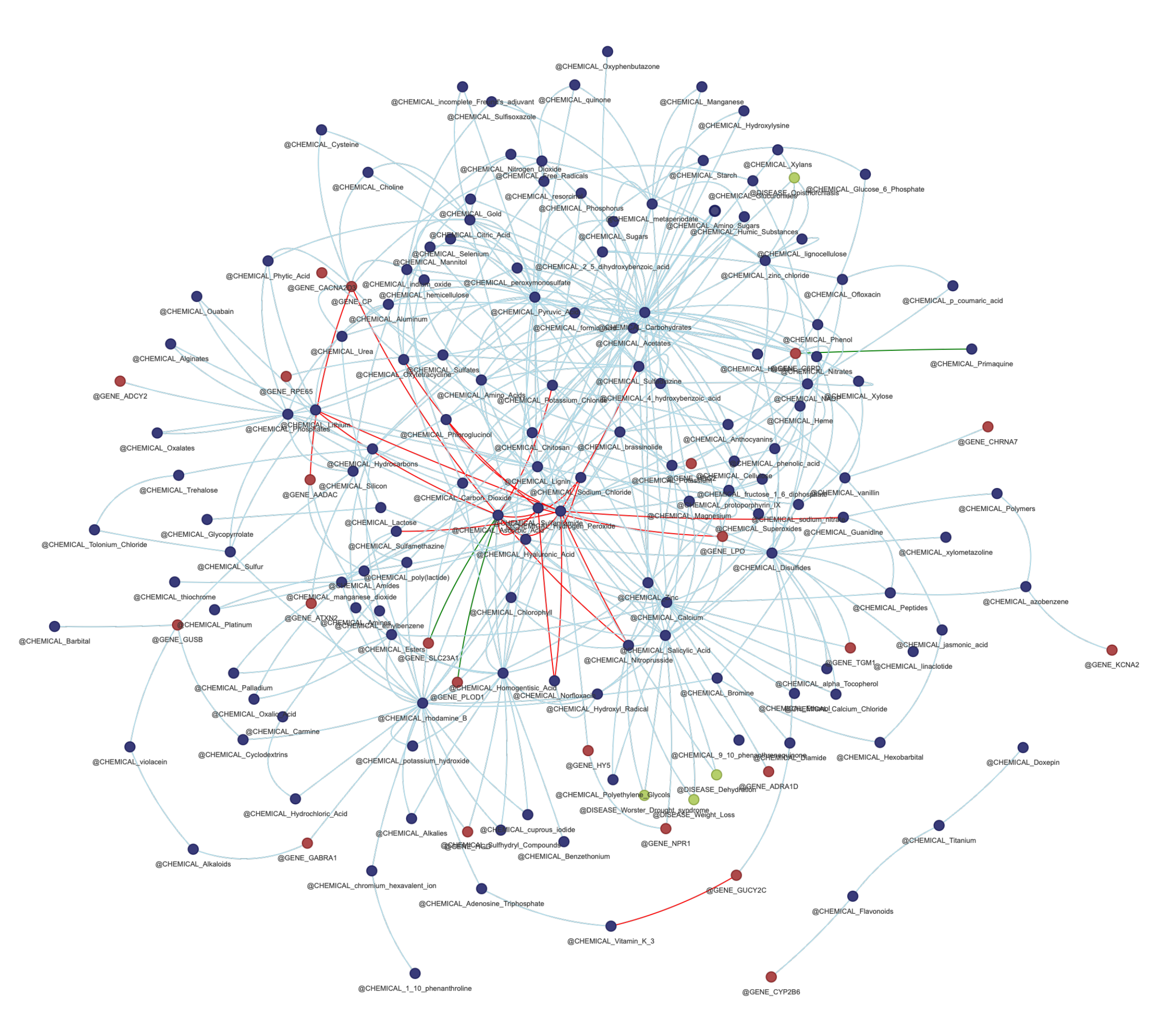}
        \caption{Subgraph constructed from the shortest paths in the \textit{Alkaptonuria} module of the \textit{extended network} corresponding to the edges from the \textit{DGIdb} graph shown in Figure~\ref{fig:exdgidbgraph}. An HTML version is available at \href{https://giangpth.github.io/Alkaptonuria/visualizations/exdgidbpaths.html}{link}.}
        \label{fig:exdgidbpaths}
    \end{subfigure}

    \begin{subfigure}{.48\linewidth}
    \centering
    \includegraphics[width=.98\linewidth]{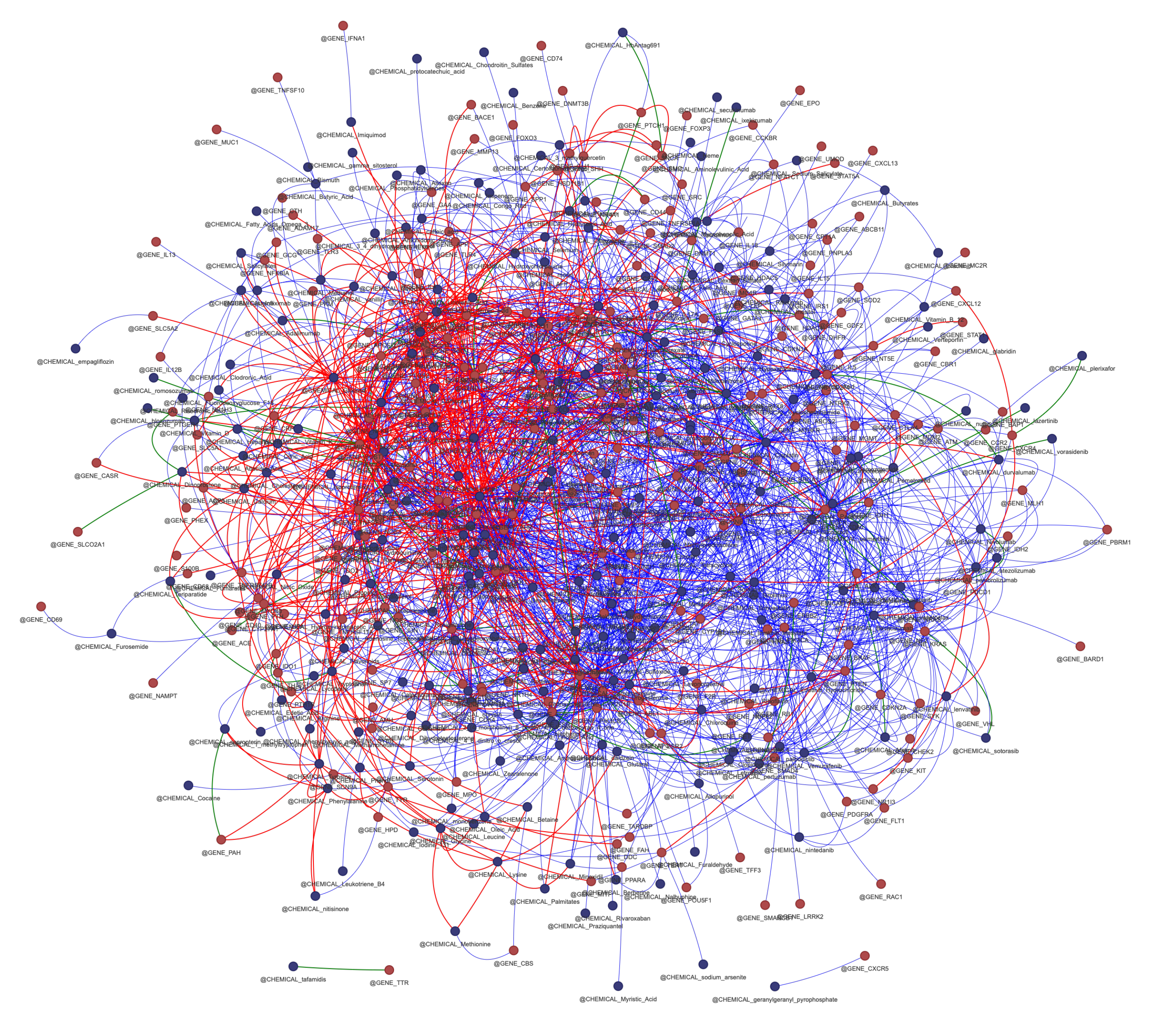}
        \caption{Visualization of gene–drug connections obtained from \textit{DGIdb} (green and blue) and from the \textit{high-confidence network} (green and red). An HTML version is available at \href{https://giangpth.github.io/Alkaptonuria/visualizations/hidgidb.html}{link}.}
        
        \label{fig:hidgidbgraph}
    \end{subfigure}
    \hfill
    \begin{subfigure}{.48\linewidth}
    \centering
    \includegraphics[width=.98\linewidth]{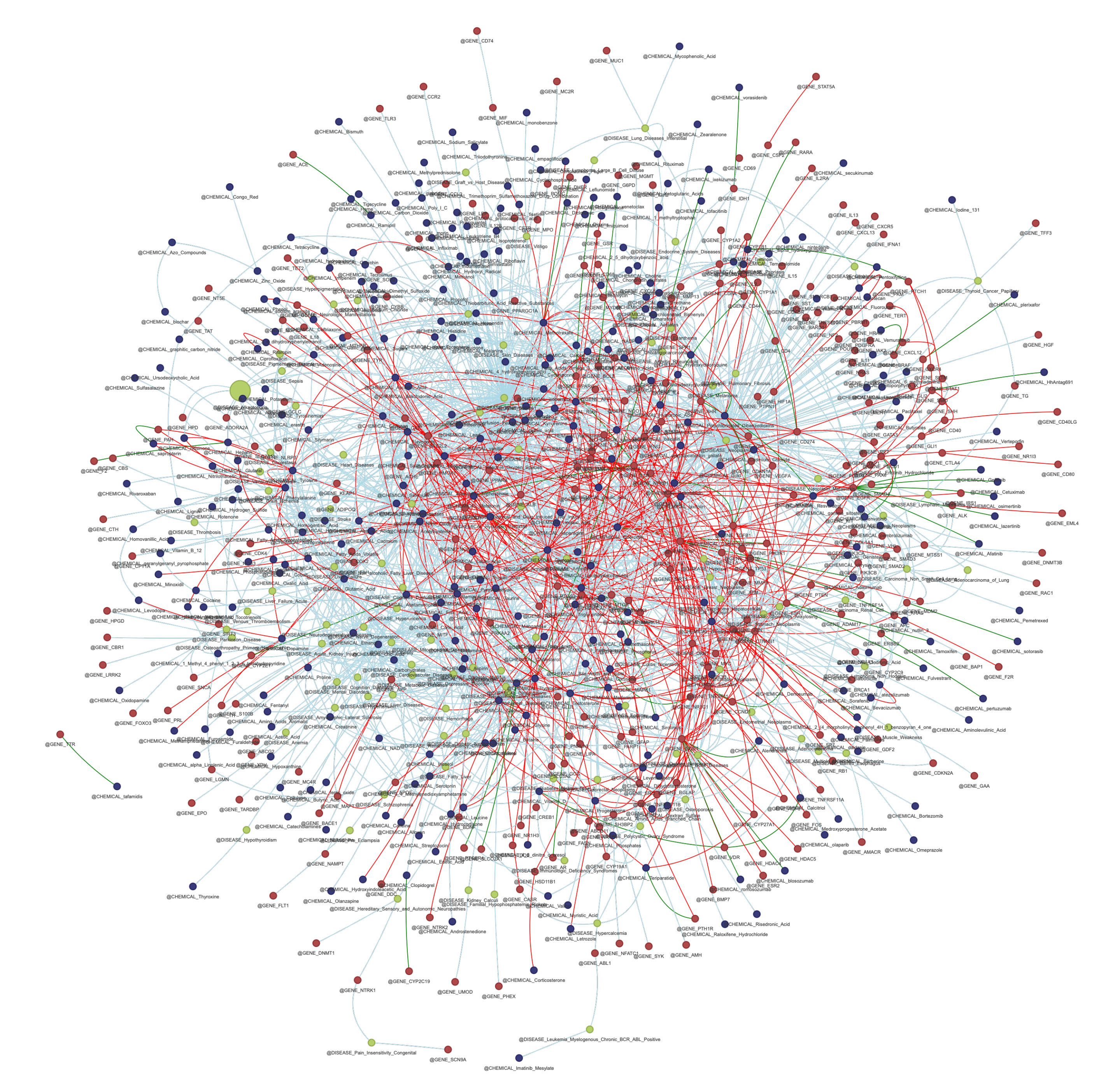}
        \caption{Subgraph constructed from the shortest paths in the \textit{high-confidence network} corresponding to the edges from the \textit{DGIdb} graph shown in Figure~\ref{fig:hidgidbgraph}. An HTML version is available at \href{https://giangpth.github.io/Alkaptonuria/visualizations/hidgidbpaths.html}{link}.}
        \label{fig:hidgidbpaths}
    \end{subfigure}
   
    \caption{Comparison between our KGs and gene–drug connections from \textit{DGIdb}.}
    \label{fig:dgidb}
\end{figure}

To validate drug-gene interactions, we display overlaps between our KGs and DGIdb in Figure~\ref{fig:dgidb}. Specifically, Figure~\ref{fig:exdgidbgraph} illustrates the DGIdb-derived network with genes from the AKU module in the extended network. Again, green edges represent interactions shared with the \textit{extended network}, blue edges denote connections absent from the \textit{extended network} but whose endpoints exist within it, red edges correspond to interactions found only in the \textit{extended network}. In this comparison, 20 red edges (33.9\%), 3 green edges (5.1\%), and 36 blue edges (61\%) were identified. When transforming direct links into shortest paths, the average length of these paths is 3.28. This is an indication that pharmacological drugs likely exert an influence on AKU-associated genes in the \textit{extended network} through intermediary (metabolic) regulators or signaling proteins, rather than through direct interactions.  This mediated influence is evidenced by the interaction between chemical-stress inducers like Hydrogen Peroxide and hub genes like G6PD. The presence of these novel red edges suggests that HGA accumulation triggers a complex signaling network involving key controllers of inflammation and oxidative stress. Consequently, red edges can serve as a hypothesis for drug targets that address the systemic signaling of AKU rather than just the primary enzymatic defect. Later in the Results section, we will further explore these edges. 

The same analysis was performed for the \textit{high-confidence network}, and the results are presented in Figure~\ref{fig:hidgidbgraph}.  We obtain 573 red edges (24.4\%), 68 green edges (2.9\%), and 1,705 blue edges (72.7\%). The average length of the paths in Figure~\ref{fig:hidgidbpaths} corresponding to direct links in DGIdb is 3.95. The blue edges and longer average path length in the \textit{high-confidence network} may be due to the more rigorous literature screening, incorporating only relations that appear in multiple publications. This suggests that experimentally validated drug-gene interactions are strongly interconnected in multi-step biological modules. The large number of red edges demonstrates that this method successfully identifies non-trivial and coherent pharmacological associations that standard curated systems such as DGIdb cannot.

\begin{figure}[h!] 
  \centering
  \begin{subfigure}{.46\linewidth}
    \centering
    \includegraphics[width=.98\linewidth]{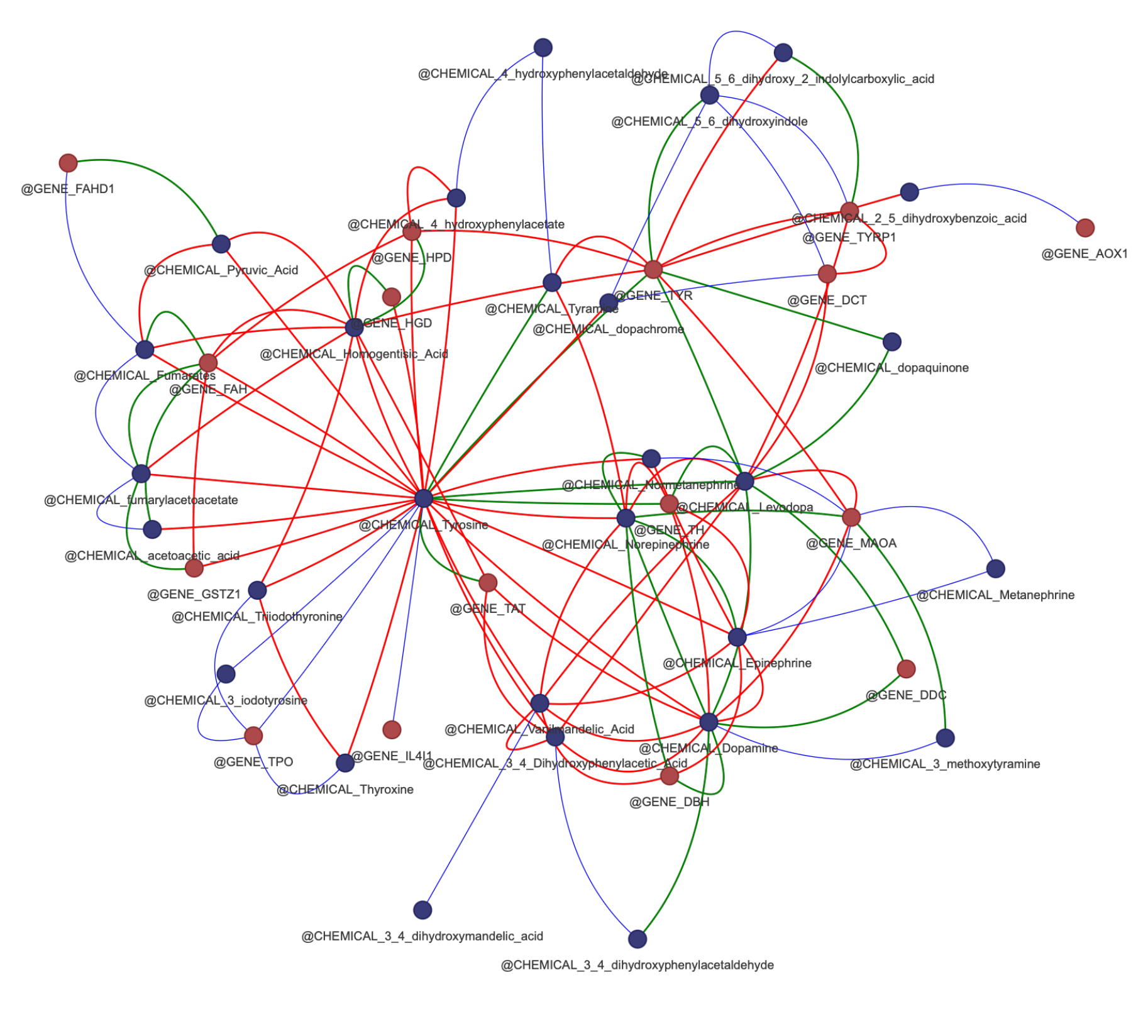}
    \caption{Visualization of connections derived from the tyrosine metabolism pathway in \textit{KEGG} (green and blue) and from the \textit{extended network} (green and red). An HTML version is available at \href{https://giangpth.github.io/Alkaptonuria/visualizations/keggexgraph.html}{link}.}
    \label{fig:keggexgrap}
  \end{subfigure}
  \begin{subfigure}{.47\linewidth}
    \centering
    \includegraphics[width=.95\linewidth]{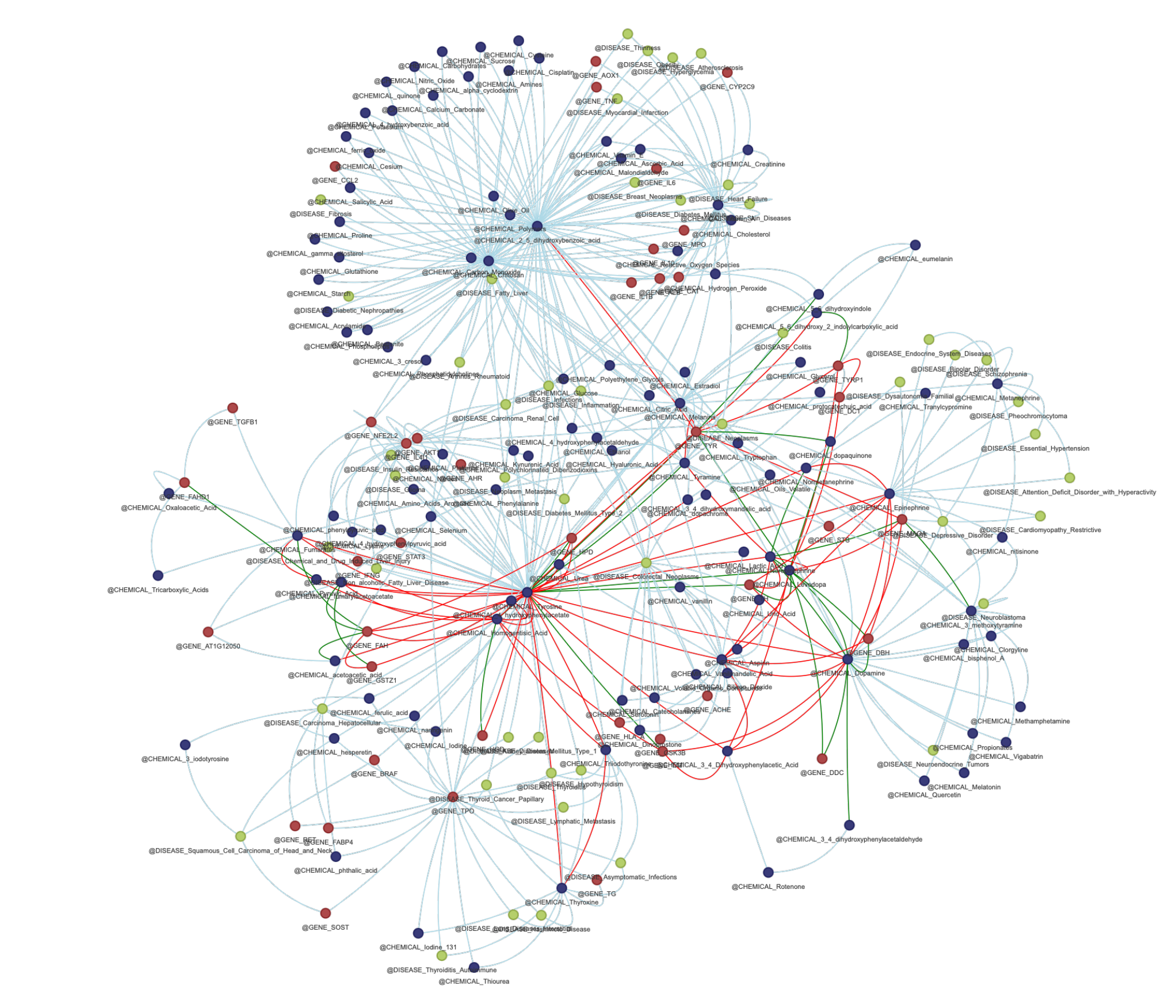}
    \caption{
    Subgraph constructed from the shortest paths in the \textit{extended network} corresponding to the edges in the tyrosine metabolism pathway. An HTML version is available at \href{https://giangpth.github.io/Alkaptonuria/visualizations/keggexpaths.html}{link}.}
    \label{fig:keggexpaths}
  \end{subfigure}
  \begin{subfigure}{.47\linewidth}
    \centering
    \includegraphics[width=.95\linewidth]{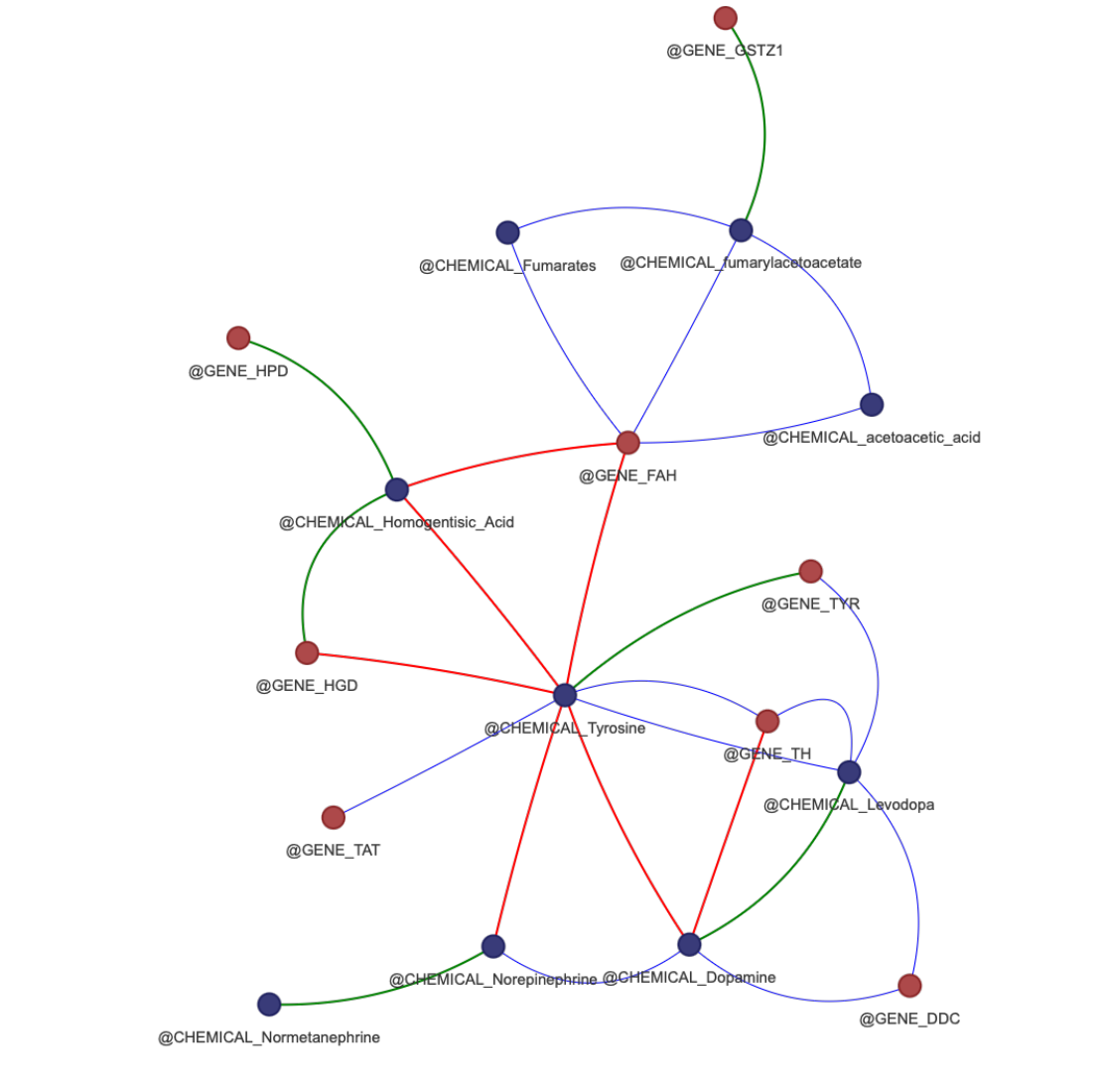}
    \caption{Visualization of connections derived from the tyrosine metabolism pathway in \textit{KEGG} (green and blue) and from the \textit{high-confidence network} (green and red). An HTML version is available at \href{https://giangpth.github.io/Alkaptonuria/visualizations/kegghigraph.html}{link}.}
    \label{fig:kegghigraph}
  \end{subfigure}
  \begin{subfigure}{.47\linewidth}
    \centering
    \includegraphics[width=.95\linewidth]{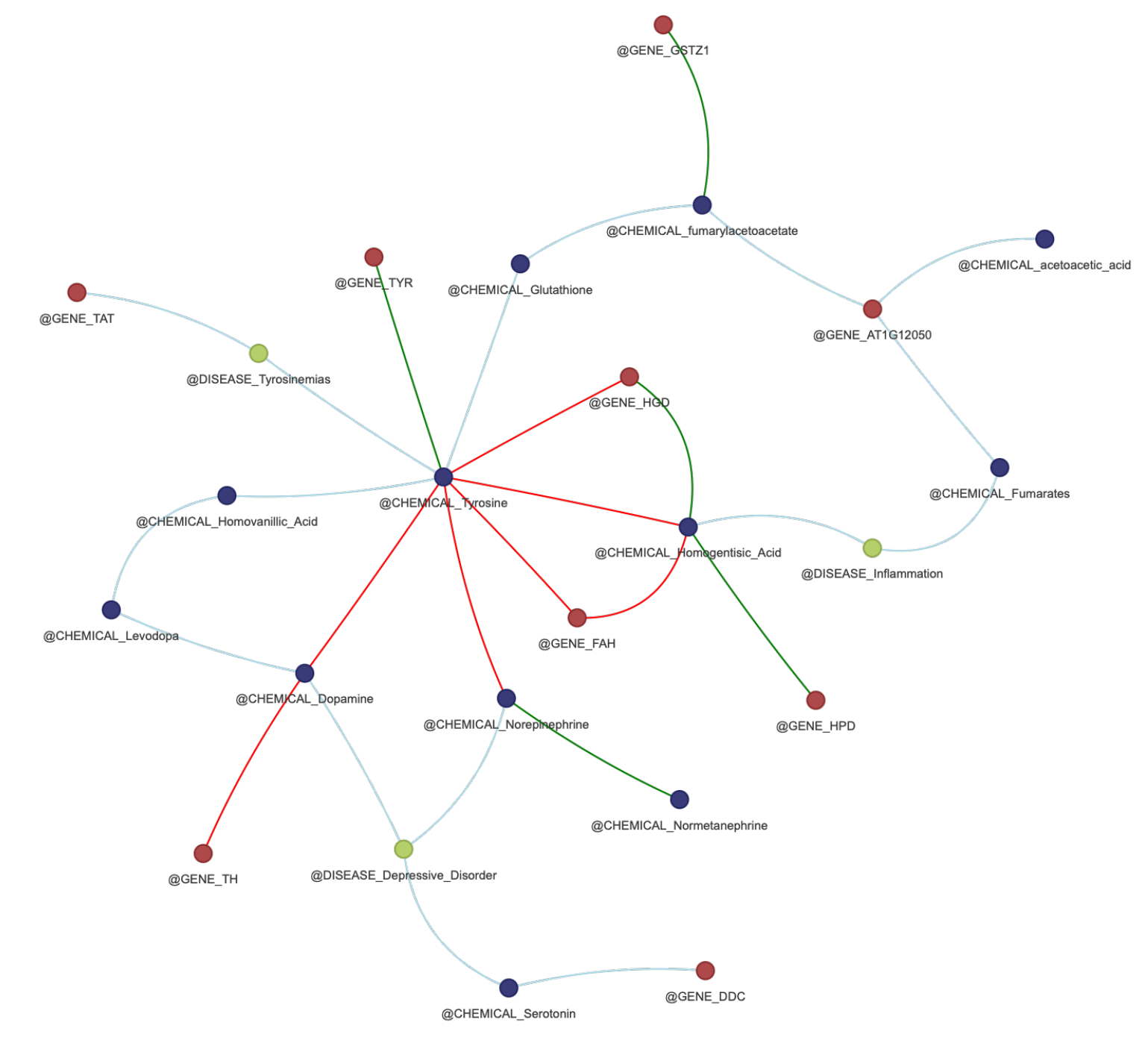}
    \caption{
    Subgraph constructed from the shortest paths in the \textit{high-confidence network} corresponding to the edges in the tyrosine metabolism pathway. An HTML version is available at \href{https://giangpth.github.io/Alkaptonuria/visualizations/kegghipaths.html}{link}.}
    \label{fig:kegghipaths}
  \end{subfigure}

  \caption{Comparison between the KGs and the network derived from the tyrosine metabolism pathway in \textit{KEGG}.}
  \label{fig:kegg}
\end{figure}

As a final validation step, we compared our Alkaptonuria KGs to  the human tyrosine metabolism pathway curated in \textit{KEGG}\cite{kanehisa2000kegg}(https://www.kegg.jp/entry/hsa00350). Figure~\ref{fig:keggexgrap} presents a comparative visualization between the extended network and the pathway. 
In this comparison, 61 red edges (52.5\%), 30 green edges (25.9\%), and 25 blue edges (21.6\%) were identified. We note that the percentage of green edges is significantly increased compared to gene-gene or gene-disease interactions, indicating that our KG has a structure more similar to a pathway.  Figure~\ref{fig:keggexpaths} further illustrates the shortest paths in the \textit{extended network} that correspond to the blue edges shown in Figure~\ref{fig:keggexgrap}. The average length of these paths is 2.56, meaning that direct blue connections in KEGG correspond on average to 1 to 2 intermediate steps in the extended network,  inferior to previous comparisons, indicating again a better representation of the pathway by our KG. The KG demonstrates a high level of specificity in recreating the enzymatic backbone of the tyrosine catabolic pathway. The network identifies the sequential conversion steps from TAT (tyrosine aminotransferase) to HPD (4-hydroxyphenylpyruvate dioxygenase), and finally to the central, disease-defining enzyme HGD. The extraction also captures the downstream enzymes GSTZ1 (maleylacetoacetate isomerase) and FAH (fumarylacetoacetate hydrolase), which represent the final stages of the fumarylacetoacetate pathway. The presence of these nodes as a coherent functional module, supported by a significant percentage of green edges, validates the methodology's ability to maintain the biological integrity of curated metabolic maps. However, the red edges in the \textit{extended network} reveal a new layer of biological information absent from traditional metabolic repositories, such as KEGG. As also previously noted in Figure~\ref{fig:exdgidbgraph}, the KG links the primary metabolic defect in HGD to high-degree hub genes associated with chronic inflammation and oxidative stress. These connections imply that the pathological accumulation of HGA activates a signaling cascade involving key regulators of cellular damage. The presence of inflammatory and stress-related genes (e.g., IL1B) together with core metabolic enzymes provides a computational hypothesis for the systemic manifestations of the disease, including premature spondyloarthropathy and cardiovascular complications. 

The same validation process was applied to the \textit{high-confidence network}, and the results are presented in Figures~\ref{fig:kegghigraph} and \ref{fig:kegghipaths}. We obtained 7 red edges (25.9\%), 6 green edges (22.2\%), and 14 blue edges (51.9\%), with an average path length of 2.79, confirming that the high-confidence network is also similar to the KEGG pathway. The validation of the \textit{high-confidence network} against the KEGG pathway confirms that stringent evidence filtering retains the indispensable metabolic core of AKU. Despite the reduction, the network successfully preserves the primary enzymatic axis of the tyrosine catabolic pathway, specifically the functional sequence involving HPD, HGD, and GSTZ1. The presence of 22.2\% green edges and a short average path length of 2.79 for blue edges shows that the high-confidence KG does not lose the connectivity of the homogentisate and fumarylacetoacetate pathways. This structural convergence ensures that the high-confidence KG provides a reliable and self-consistent biochemical module.

\subsection*{Knowledge extraction}

\subsubsection*{Network Characterization}
The \textit {extended network} exhibited a diameter of 10, indicating a tightly connected network. We identified the most centrally located nodes, in the sense that they are closest, on average, in the worst case, to all other nodes. These included molecules, genes, and diseases for which the common themes are oxidative stress, inflammation, and cellular metabolism. Among the genes that are regulators of inflammation, oxidative stress, apoptosis, and cell proliferation, there were: TP53, CTNNB1, TNF, IL6, IL1B, VEGFA, AKT1, IL10, AHR, NFKB1, ACSL4, HMOX1, MAPK8, MAPK1, APP, SIRT1, GPX4. The most central diseases are Neoplasms, Colorectal Neoplasms, Inflammation, Nerve Degeneration, Fibrosis, Osteoporosis, Skin Diseases, and Osteosarcoma. AKU was located at a distance of one from the central disease nodes. Forty-three chemicals were also centrally located.  In comparison, \textit {the high-confidence network} shared the same diameter but exhibited a single central node, Neoplasms, to which AKU was positioned at a distance of two. 

\begin{figure}[ht] 
    \centering
    \begin{subfigure}[t]{0.48\textwidth}
        \centering
        \includegraphics[width=\textwidth]{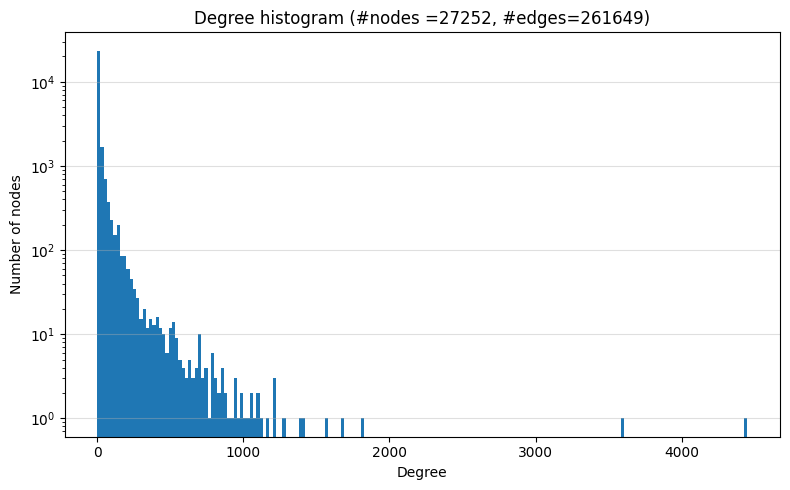}
        \caption{Histogram of node degree distribution in the \textit{extended network}.}
        \label{fig:deghis}
    \end{subfigure}
    \hfill
    \begin{subfigure}[t]{0.48\textwidth}
        \centering
        \includegraphics[width=\textwidth]{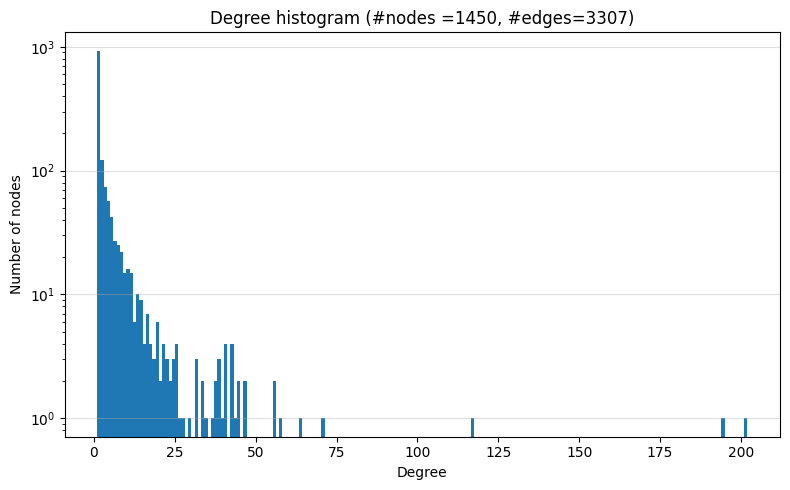}
        \caption{Histogram of node degree distribution in the \textit{high-confidence network}.}
        \label{fig:hideghis}
    \end{subfigure}

    \begin{subfigure}[t]{0.48\textwidth}
        \centering
        \includegraphics[width=\textwidth]{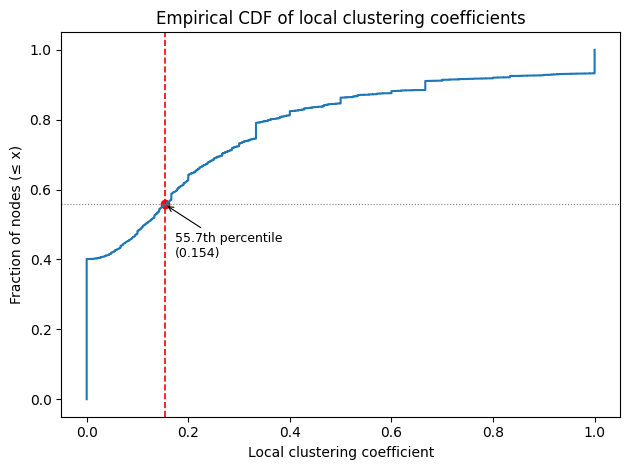}
        \caption{The distribution of the clustering coefficients of nodes in the extended network.}
        \label{fig:clustercoff}
    \end{subfigure}
    \hfill
    \begin{subfigure}[t]{0.48\textwidth}
        \centering
        \includegraphics[width=\textwidth]{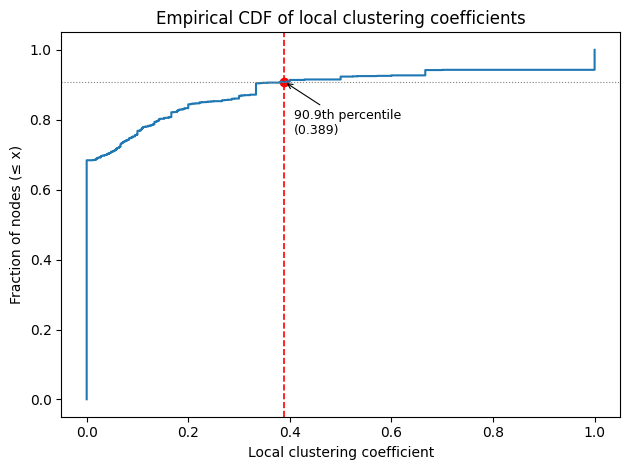}
        \caption{The distribution of the clustering coefficients of nodes in the high-confidence network}
        \label{fig:hiclustercoff}
    \end{subfigure}
    
    \caption{Distribution of node degrees and clustering coefficients}
    \label{fig:characterisation}
\end{figure}

Regarding node connectivity, Figure \ref{fig:characterisation} shows the degree distributions of the two networks, illustrating how nodes are distributed according to their number of connections. In both cases, we note long-tailed distributions typical of biological networks, where many nodes have few connections, but some nodes are important hubs. The figure also presents the distributions of local clustering coefficients. The extended network displays an average clustering coefficient of 0.2184 and a transitivity of 0.0731, indicating moderate local connectivity. In contrast, the high-confidence network demonstrates lower overall cohesion, with an average clustering coefficient of 0.114 and a transitivity of 0.0576. Notably, the node representing AKU (marked by a red dashed line) exhibits a higher local clustering coefficient in the high-confidence network (0.389; 90th percentile) compared with the extended network (0.154; 50th percentile), suggesting that, as expected, AKU is embedded within a more tightly connected neighborhood when only high-confidence associations are considered. 

The degree correlation is negative in both networks, with values of $-0.12$ for the extended network and $-0.137$ for the high-confidence network. This indicates a disassortative structure in which high-degree nodes function as hubs that preferentially connect to low-degree nodes.

\subsubsection*{Community detection}
We study the communities present in the two KGs to uncover the modular organization of entities and possible relationships. We employ three different techniques for detecting communities: modularity-based (Leiden algorithm), k-core analysis and clique analysis (see Methods for more details). We concentrate on the communities where AKU belongs.

\begin{figure}[] 
\centering
\begin{subfigure}[t]{0.7\textwidth}
    \centering
    \includegraphics[width=.85\linewidth]{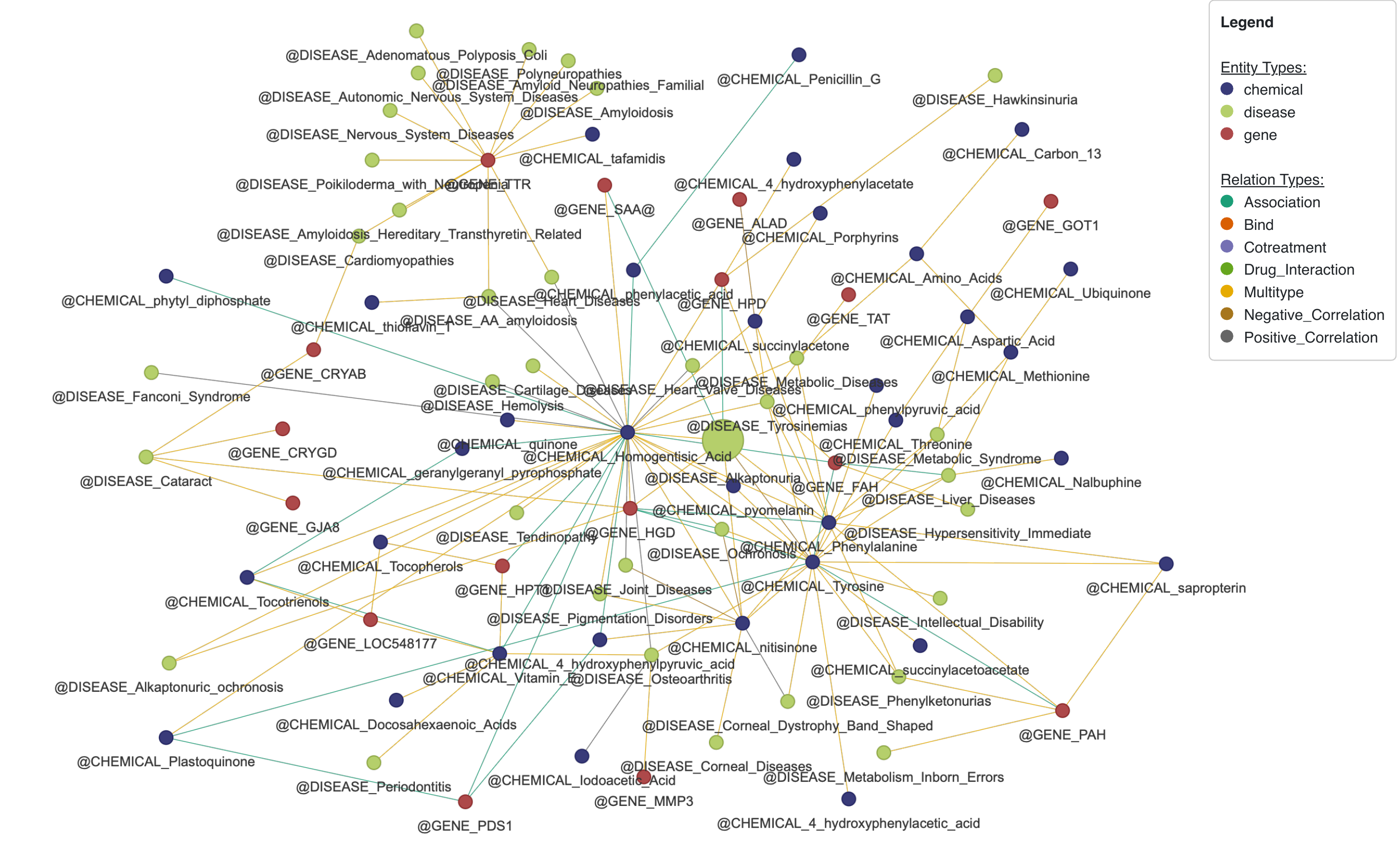}\caption{Module containing Alkaptonuria in the \textit{high-confidence network, based on the Leiden algorithm. An HTML view of the network is available at the \href{https://giangpth.github.io/Alkaptonuria/visualizations/himodule.html}{link}.  }}
    \label{fig:himodule}
\end{subfigure}  
\\
 \begin{subfigure}[t]{0.48\textwidth}
        \centering
        \includegraphics[width=\textwidth]{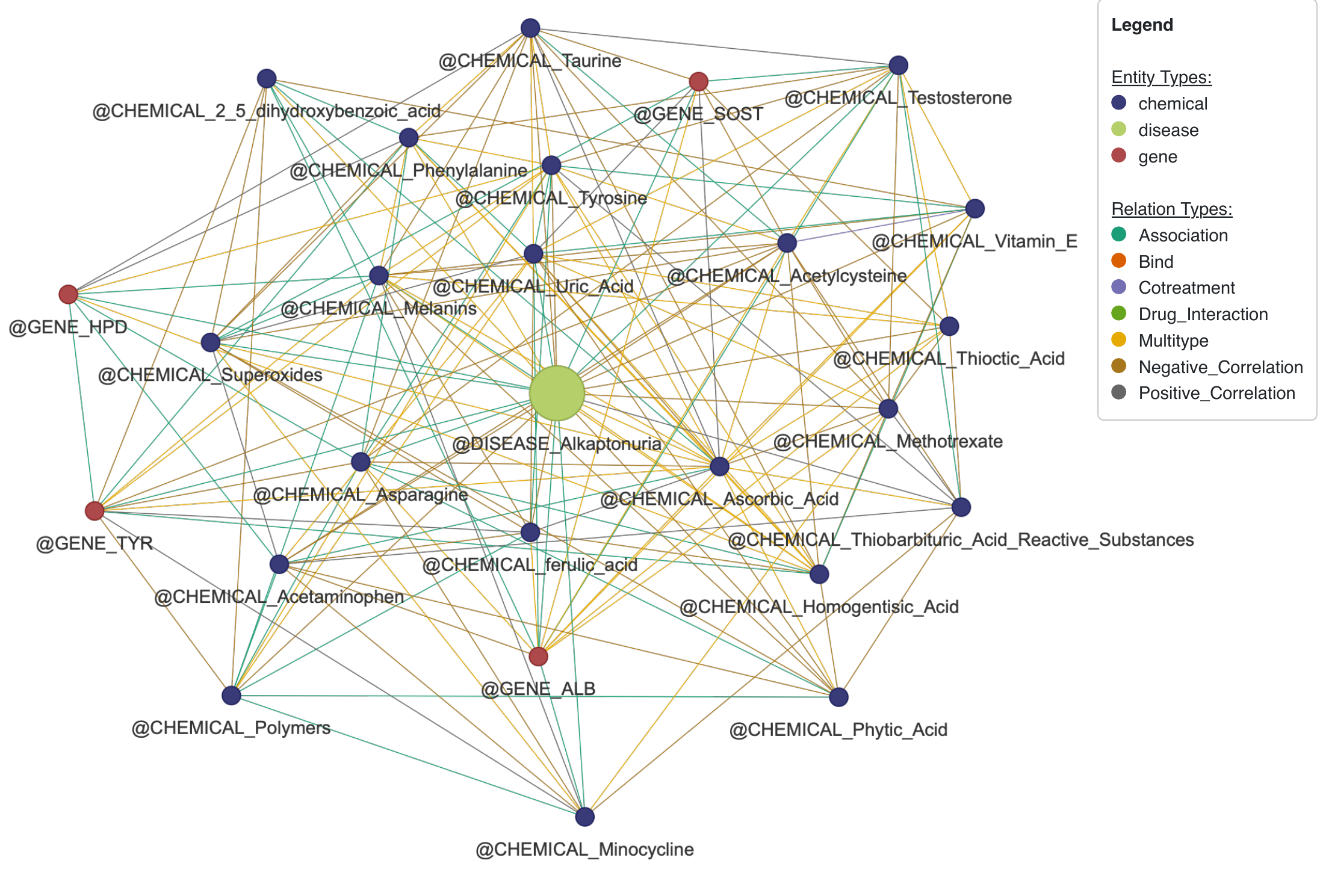}
        \caption{Maximum k-core subgraph with k=9 of Alkaptonuria in the extended network. An HTML view of the network is available at the \href{https://giangpth.github.io/Alkaptonuria/visualizations/excore.html}{link}.}
        \label{fig:excore}
    \end{subfigure}
    \begin{subfigure}[t]{0.48\textwidth}
        \centering
        \includegraphics[width=\textwidth]{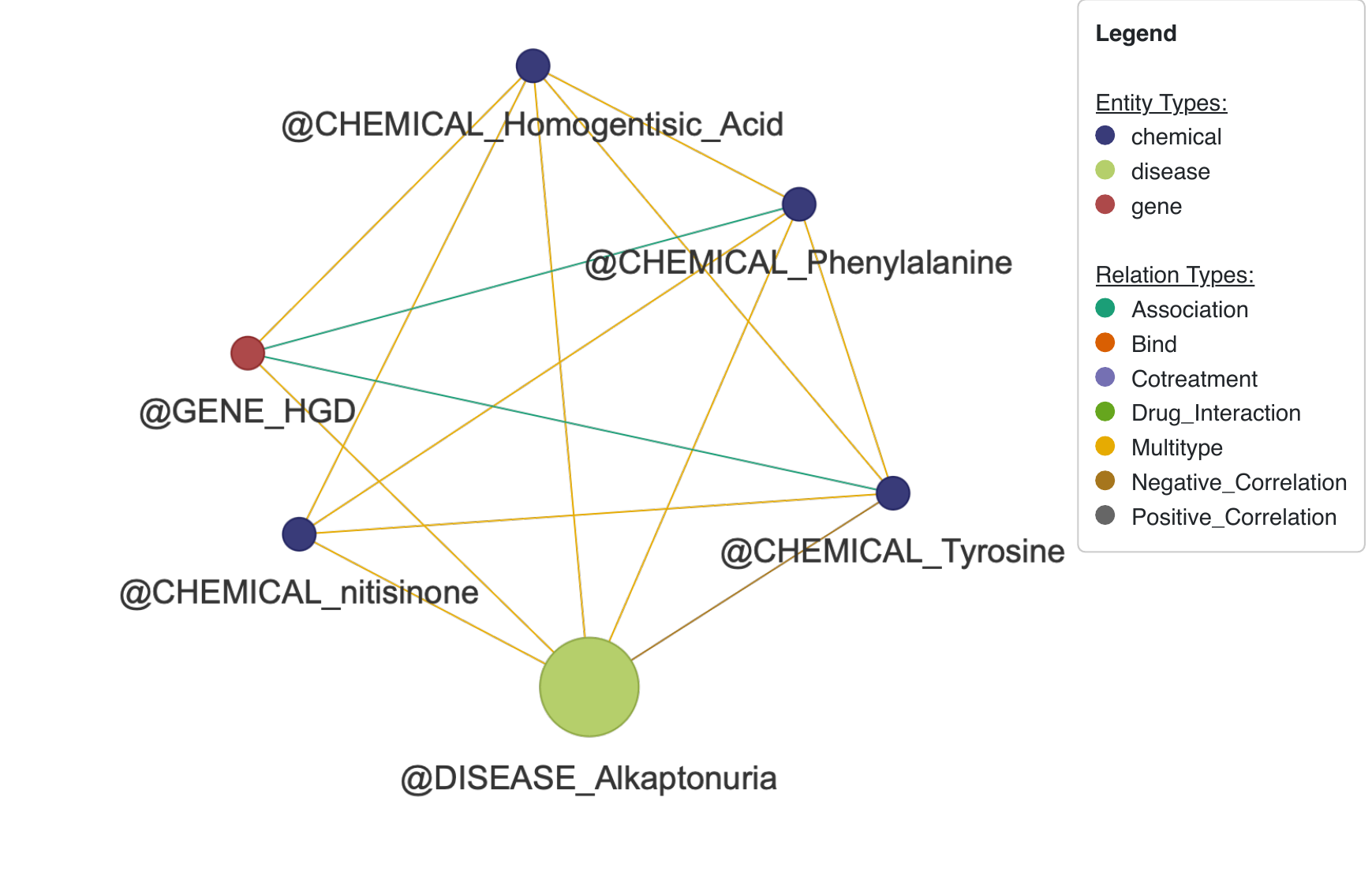}
        \caption{Maximum k-core subgraph with k=4 of Alkaptonuria in the high-confidence network. An HTML view of the network is available at the \href{https://giangpth.github.io/Alkaptonuria/visualizations/hicore.html}{link}.}
        \label{fig:hicore}
    \end{subfigure}
    \begin{subfigure}[t]{0.48\textwidth}
        \centering
        \includegraphics[width=\textwidth]{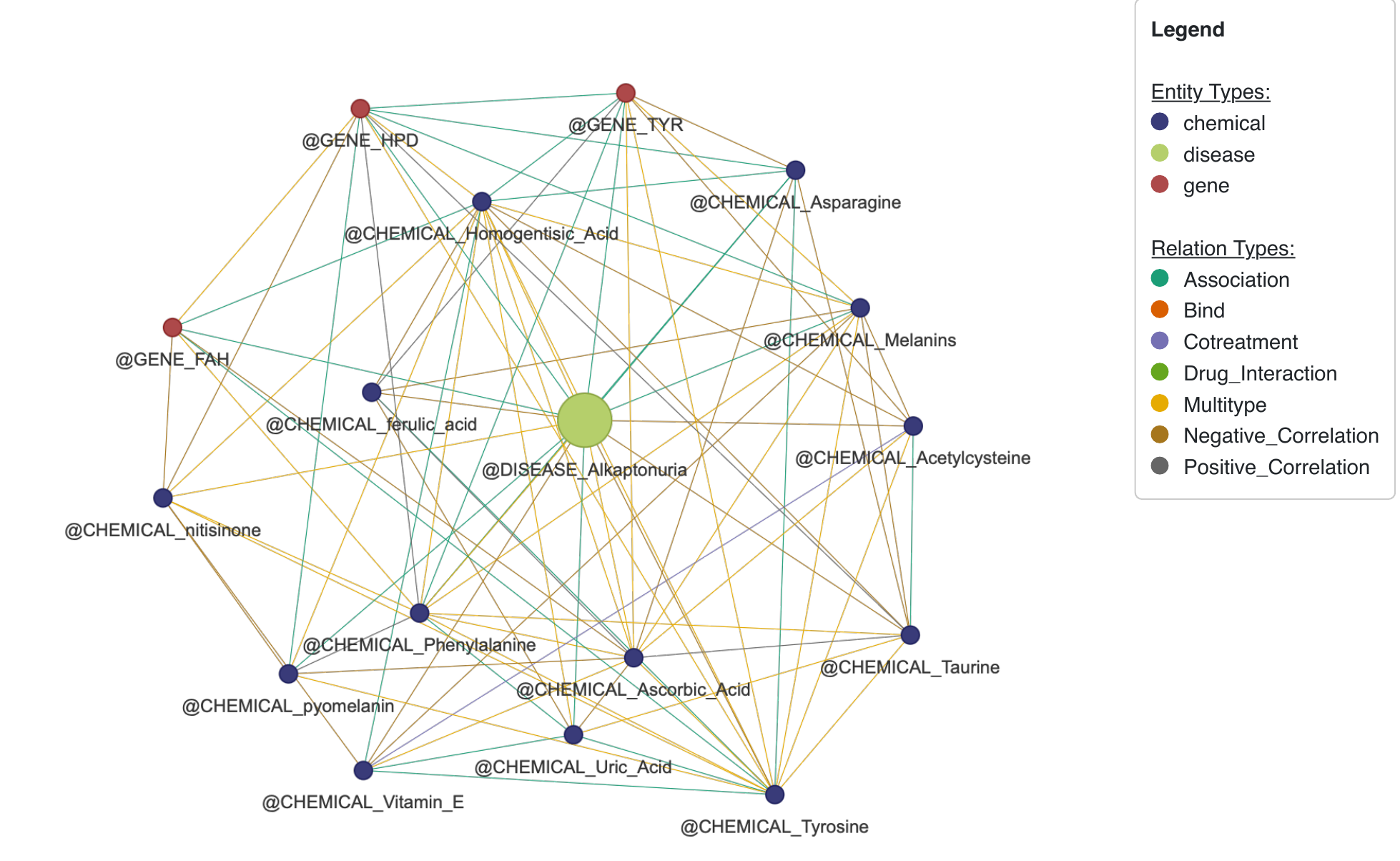}
        \caption{Subgraph from the \textit{extended network} containing nodes belonging to 15 maximum cliques (of size 7) associated with {Alkaptonuria}. An HTML view of the network is available at the \href{https://giangpth.github.io/Alkaptonuria/visualizations/maxcliques.html}{link}.}
        \label{fig:clique}
    \end{subfigure}
    \hfill
    \begin{subfigure}[t]{0.48\textwidth}
        \centering
        \includegraphics[width=\textwidth]{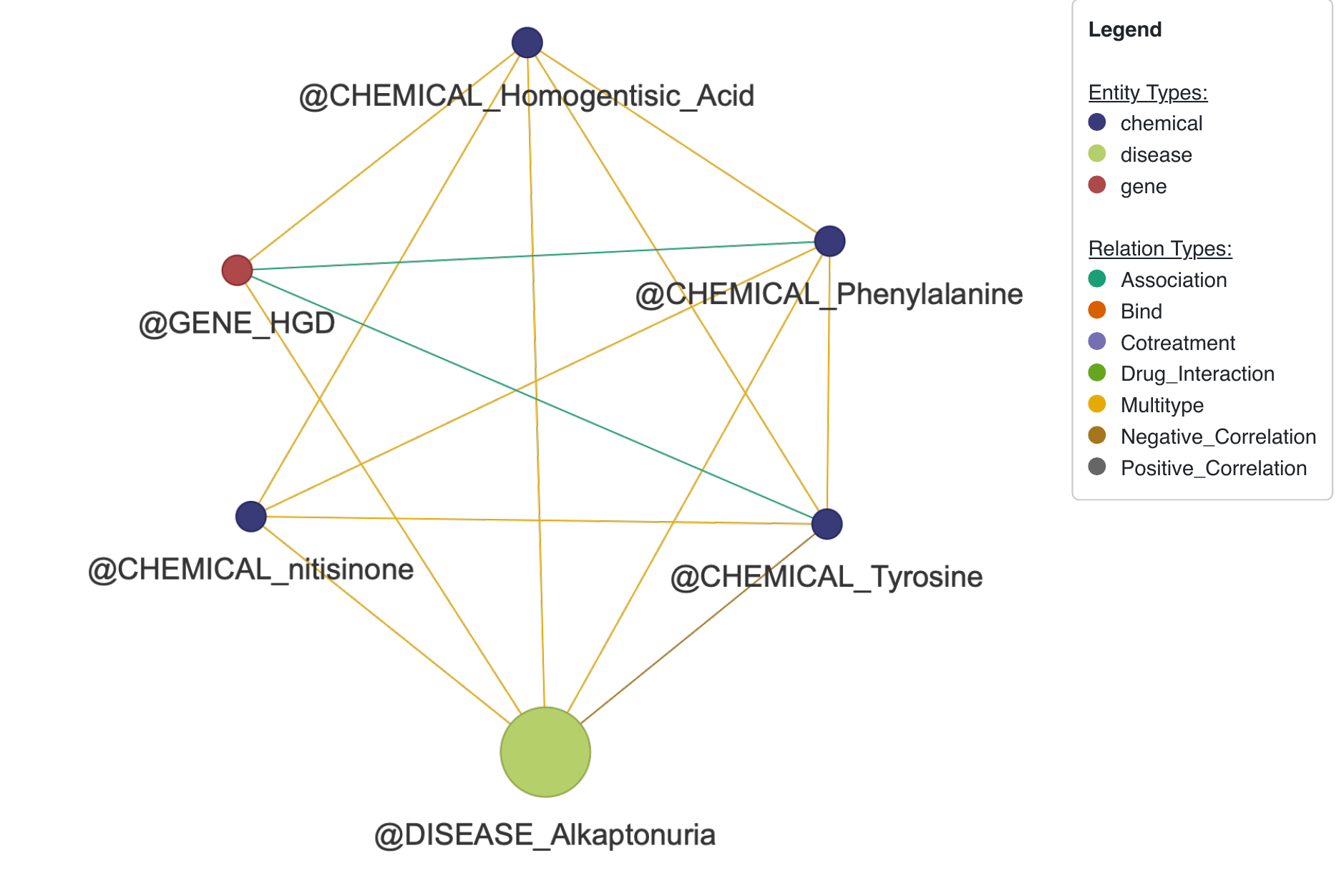}
        \caption{Subgraph from the \textit{high-confidence network} containing nodes belonging to 2 maximum cliques (of size 5) associated with {Alkaptonuria}. An HTML view of the network is available at the \href{https://giangpth.github.io/Alkaptonuria/visualizations/himaxcliques.html}{link}.}
        \label{fig:hiclique}
    \end{subfigure}
    \caption{Community detection. AKU communities extracted by different community detection approaches.}
\end{figure}

Based on the Leiden algorithm, the \textit{extended network} consists of 34 modularity-based communities, with a modularity score of $Q = 0.336$.  The \textit{high-confidence network} comprises 17 communities with a higher modularity score of $Q = 0.546$. Both scores fall within the typical range of $0.3$ to $0.7$ observed in natural networks~\cite{newman2004finding}. The higher score of the high-confidence network indicates a tighter community structure with fewer links across communities, while the extended network appears to be more uniformly connected, as also noted above with clustering coefficients.  In the \textit{extended network}, the disease {AKU}, together with {HGD} and {Homogentisic acid}, are located in the second-largest community containing 4,008 nodes. Meanwhile, in the \textit{high-confidence network}, they belong to a module of 84 nodes. Figure~\ref{fig:himodule} displays the module containing {Alkaptonuria} in the \textit{high-confidence network}. We do not display the extended network module due to its size. The AKU-centered module in the \textit{high-confidence network} reflects a condensed version of the biochemical–pathophysiological core tightly associated with HGA metabolism and its systemic consequences. AKU, HGA, and HGD connect to a set of metabolic intermediates, including tyrosine- and phenylalanine-related compounds, and to clinically relevant disease phenotypes. An interesting link is to the TTR gene, which appears to be central to a cluster containing also AA amyloidosis, a well-established comorbidity of AKU driven by chronic inflammation. The presence of amyloidosis-related nodes within this compact module supports the view that amyloid deposition is an intrinsic component of AKU pathophysiology rather than a secondary, unrelated complication \cite{millucci2012alkaptonuria}. The two diseases could share similar pathways, and the centrality of TTR could be investigated further. Associations with liver disorders, inborn errors of metabolism, and connective tissue diseases further reflect the multisystemic nature of AKU, in which HGA accumulates systemically and undergoes oxidative polymerization. The links to cartilage degeneration, osteoarticular disease, and cardiovascular disorders highlight the central role of ochronotic pigment deposition in collagen-rich tissues and its contribution to chronic inflammation and tissue damage, processes known to promote AA amyloidogenesis\cite{Millucci2014}. Another interesting set of connections are those between nitisinone, tyrosine and the many diseases associated with these nodes, including neurological disorders. These links could drive the exploration of possible side effects of nitisinone treatment. Furthermore, we note that our KG connects AKU to disorders like tyrosinemia, FAH-related diseases and  HPD-related metabolism, thus involving the entire aromatic amino acid degradation pathway. This underlines the fact that AKU is part of a \emph{disease family} of metabolic blocks in the same pathway.

We also display the maximum k-core subgraphs containing \textit{Alkaptonuria} in both the \textit{extended network} (Figure~\ref{fig:excore}) and the \textit{high-confidence network} (Figure~\ref{fig:hicore}). The maximum core index for \textit{Alkaptonuria} in the \textit{extended network} is 9, corresponding to a core size of 26 nodes, whereas in the \textit{high-confidence network} the core index is 4 with a core size of 6 nodes. The k-core structures highlight a clear difference in the biological scope captured by the two networks. In the \textit{extended network} (Figure~\ref{fig:excore}), the AKU-centered core incorporates a heterogeneous set of metabolites and genes, reflecting downstream and systemic processes that accompany disease progression. Notably, no diseases besides AKU are present in this module, therefore it displays a molecular neighborhood of AKU. The graph suggests that outside the core AKU module, the most important biology is the overlap between aromatic amino-acid metabolism, ochronotic pigment/polymer formation, and antioxidant/oxidative-stress chemistry. The \textit{high-confidence} core (Figure~\ref{fig:hicore}) isolates a compact biochemical module directly aligned with aromatic amino acid metabolism and its pharmacological modulation. This smaller network indicates that while the \textit{extended network }contextualizes AKU within broader systemic consequences, the \textit{high-confidence network} preferentially preserves the core metabolic mechanism underlying the disease.

As for clique analysis, in the extended network, the disease Alkaptonuria belongs to 15 maximum cliques of size 7; meanwhile, in the high-confidence network, it belongs to 2 maximum cliques of size 5. Figure \ref{fig:clique}  shows the subgraphs containing the maximum cliques including Alkaptonuria in the extended network, while that extracted from the high-confidence graph is shown in Figure~\ref{fig:hiclique}. 
In the \textit{extended network} (Figure \ref{fig:clique}), AKU participates in a larger number of maximum cliques that substantially overlap with the corresponding k-core above but it is smaller and includes some additional nodes. 
It confirms the view that AKU peripheral biology sits at the intersection of tyrosine/homogentisate metabolism, pigment-like oxidation chemistry, and antioxidant chemistry. However, it adds a clearer treatment-centered layer through nitisinone and a more specific pigment-product layer through pyomelanin, while extending the pathway downstream with FAH, towards a more complete tyrosine degradation pathway.
In the \textit{high-confidence network}, the maximum clique subgraph (Figure~\ref{fig:hiclique}) is identical to the maximum k-core (Figure~\ref{fig:hicore}), indicating that the AKU-centered core forms a fully connected and self-consistent biochemical module. This structural convergence reflects the strong and unambiguous literature support for the core metabolic mechanism underlying the disease.

\subsubsection*{Entities related to Alkaptonuria}
\label{sec:analysis}
In this section, we further explore both the \textit{extended network} and the \textit{high-confidence network} to identify the main nodes related to AKU. We use Personalized PageRank (PPR), with AKU as the personalization source node, and Personalized Katz Centrality, to rank central entities (see Methods for more details). 

\begin{figure}[] 
  \centering
  \begin{subfigure}{\linewidth}
    \centering
    \includegraphics[width=.9\linewidth]{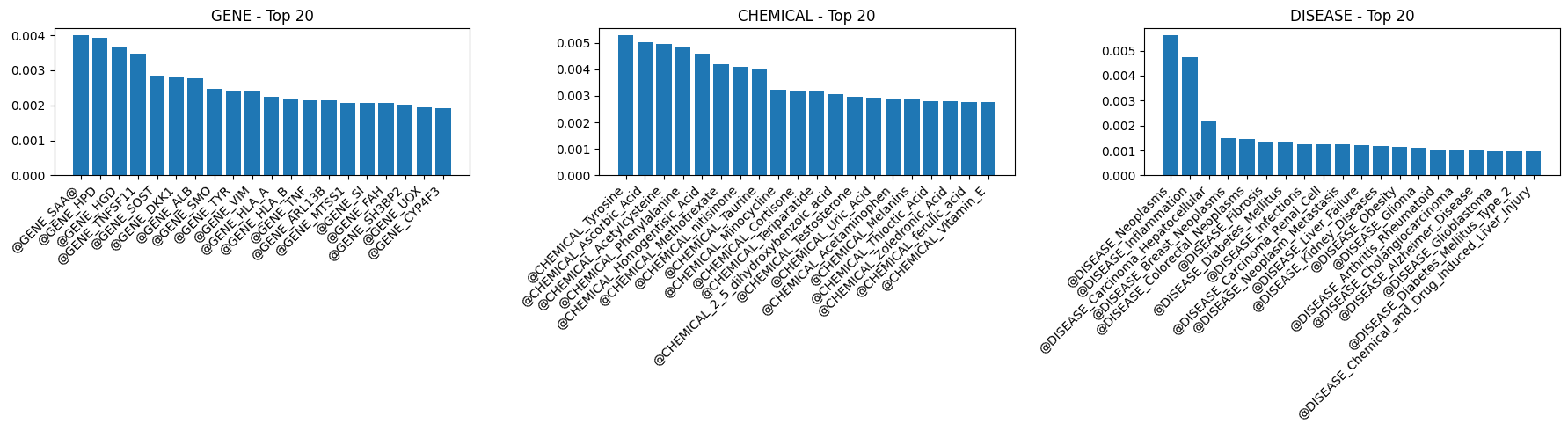}
    \caption{Top 20 nodes in each category (\textit{gene}, \textit{chemical}, \textit{disease}) ranked by PPR scores computed using \textit{Alkaptonuria} as the personalization source in the \textit{extended network}. An HTML version of the subgraph determined by these entities can be seen at \href{https://giangpth.github.io/Alkaptonuria/visualizations/exPPRsubgraph.html}{link}. }
    \label{fig:pprextop}
  \end{subfigure}

   \begin{subfigure}{\linewidth}
    \centering
    \includegraphics[width=.9\linewidth]{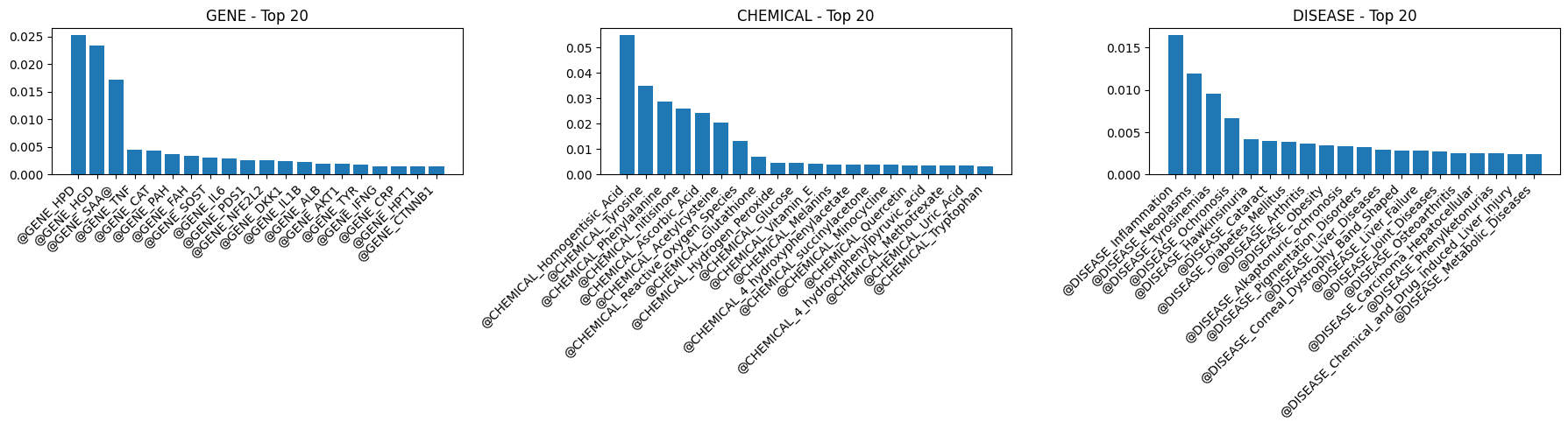}
    \caption{Top 20 nodes in each category (gene, chemical, disease) ranked by Personalized PageRank (PPR) scores computed with Alkaptonuria as the personalization source in the high-confidence network. An HTML version of the subgraph determined by these entities can be seen at \href{https://giangpth.github.io/Alkaptonuria/visualizations/hiPPRsubgraph.html}{link}.}
    \label{fig:pprhitop}
  \end{subfigure}

    \begin{subfigure}{\linewidth}
    \centering
    \includegraphics[width=.9\linewidth]{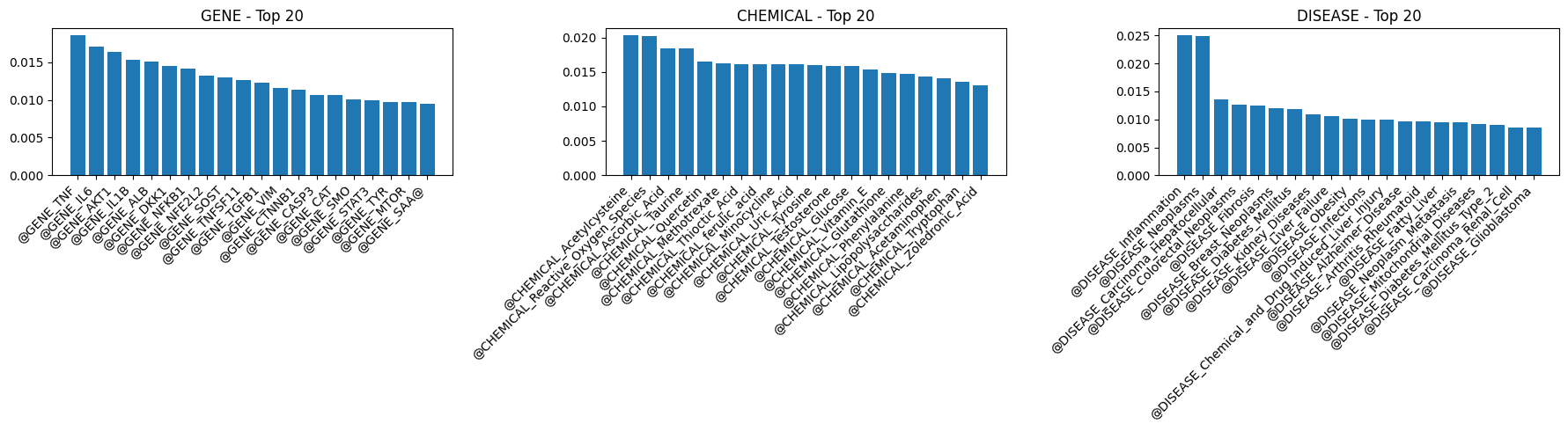}
    \caption{Top 20 nodes in each category (\textit{gene}, \textit{chemical}, \textit{disease}) ranked by Personalized Katz centrality scores computed using \textit{Alkaptonuria} as the personalization source in the \textit{extended network}. An HTML version of the subgraph determined by these entities can be seen at \href{https://giangpth.github.io/Alkaptonuria/visualizations/exKatzsubgraph.html}{link}. }
    \label{fig:exkatztop}
  \end{subfigure}

  \begin{subfigure}{\linewidth}
    \centering
    \includegraphics[width=.9\linewidth]{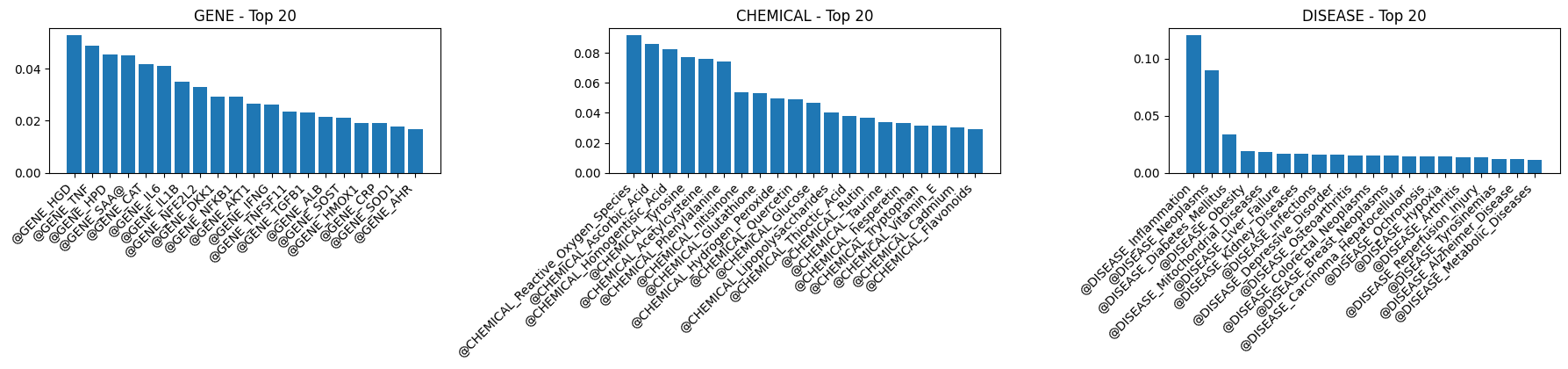}
    \caption{Top 20 nodes in each category (\textit{gene}, \textit{chemical}, \textit{disease}) ranked by Personalized Katz centrality scores computed using \textit{Alkaptonuria} as the personalization source in the \textit{high-confidence network}. An HTML version of the subgraph determined by these entities can be seen at \href{https://giangpth.github.io/Alkaptonuria/visualizations/hiKatzsubgraph.html}{link}. }
    \label{fig:hikatztop}
  \end{subfigure}

  \caption{Nodes with highest PPR and Katz centrality score from the two KGs. }
  \label{fig:centrality}
\end{figure}

Figures \ref{fig:pprextop} and \ref{fig:pprhitop} show the ranking and scores of the top 20 nodes in each category (gene, chemical, and disease) with the highest PPR scores, in the extended network and the high-confidence network, respectively. For an easier exploration of the entities, we also include links to HTML views of the subnetworks that those entities define (in the figure captions). Across both networks, the entities prioritized by PPR form a biologically coherent neighborhood centered on tyrosine metabolism and its systemic consequences. In particular, the prominence of enzymes directly involved in HGA processing, along with metabolites belonging to the aromatic amino acid degradation pathway, suggests that the ranking reflects functional proximity rather than simple network connectivity. At the disease level, the presence of inflammatory, metabolic and degenerative conditions reflects the well-documented consequences of long-term accumulation of HGA. Notably, convergence of these patterns in both the \textit{extended} and \textit{high-confidence} networks supports the robustness of the biological signals identified, indicating they are not driven by random text-mining associations.

Besides PPR,  Figures~\ref{fig:exkatztop} and~\ref{fig:hikatztop} present the top 20 nodes in each category with the highest Personalized Katz centrality scores in the \textit{extended} and \textit{high-confidence} networks, respectively. Again, the subnetworks are available as links in the figure captions. We note entities that influence AKU through longer-range functional connections rather than proximity alone. The enrichment of genes and chemicals involved in oxidative stress, signaling and metabolic regulation, together with systemic disease categories, indicates the contribution of indirect pathways to disease modulation. The consistency of these patterns in the \textit{high-confidence network} confirms the reliability of the long-range biological signals captured by this method.


 We underline some possibly interesting patterns from Figure~\ref{fig:centrality}, for future research. The inflammation axis around TNF is very prominent, suggesting a large inflammatory cytokine component. Since TNF drives chronic inflammatory degeneration of connective tissue and is central in rheumatoid arthritis and cartilage damage, ochronosis-related joint degeneration could plausibly involve TNF-driven inflammation. So the KG reinforces the idea that AKU may share inflammatory mechanisms with arthritis and chronic inflammatory diseases. A different pathway that appears is bone remodeling, with genes such as DKK1, SOST, and TNFSF11. These could hint into the mechanisms behind spinal degeneration,
joint damage and 
calcification observed in AKU patients, through bone remodeling signaling, not just pigment deposition. In terms of therapeutic insights, we note the presence of multiple antioxidant chemicals, such as Vitamin-C, Thioctic-acid, and anti-inflammatory (Vitamin-E), some of which are not yet employed in AKU management. Combined with the strong presence of oxidative and inflammatory pathways, these could indicate possible avenues for drug repurposing. 

\begin{figure}[ht]
    \centering
    \begin{subfigure}[t]{0.48\textwidth}
        \centering
        \includegraphics[width=\textwidth]{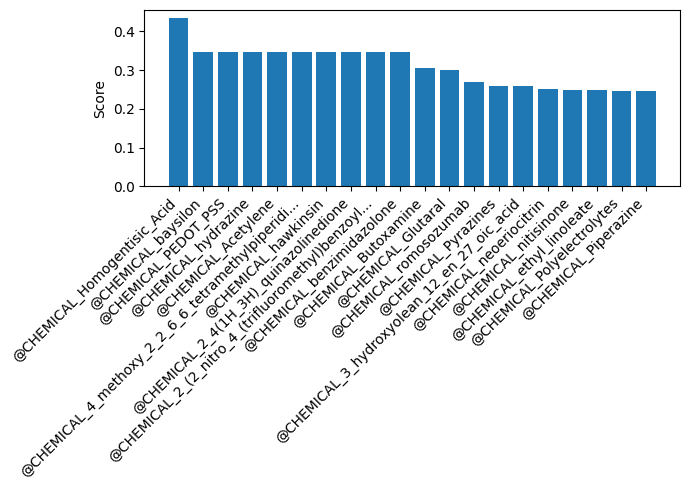}
        \caption{Top 20 chemicals with highest HeteSim score in the extended network.}
        \label{fig:hetesimex}
    \end{subfigure}
    \hfill
    \begin{subfigure}[t]{0.48\textwidth}
        \centering
        \includegraphics[width=\textwidth]{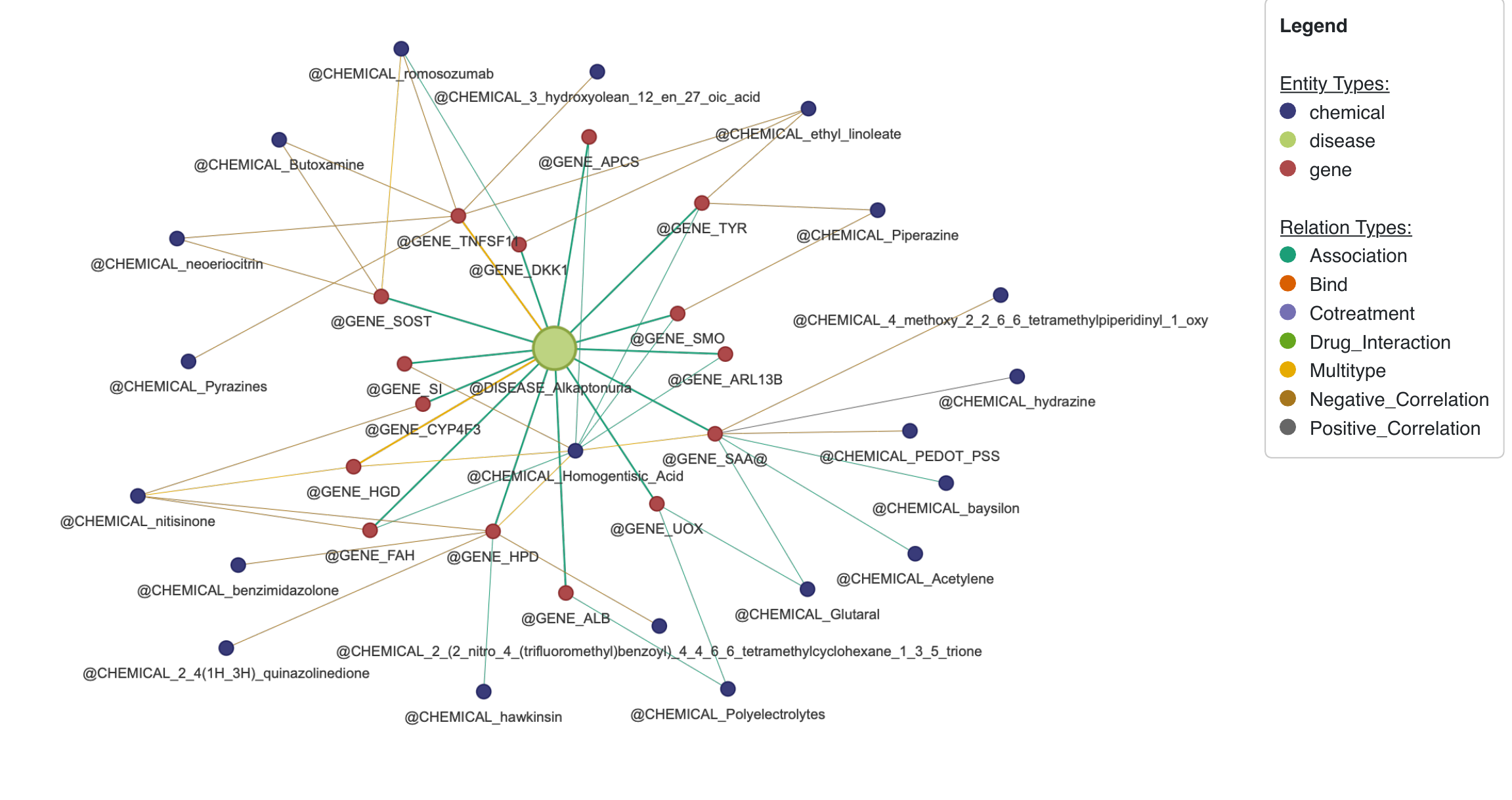}
        \caption{Meta-paths of top 20 chemicals with highest HeteSim score in the extended network. An HTML view of the meta-paths is available at \href{https://giangpth.github.io/Alkaptonuria/visualizations/exHeteSimsubgraph.html}{link}.}
        \label{fig:hetesimexgraph}
    \end{subfigure}
    \hfill

    \centering
    \begin{subfigure}[t]{0.48\textwidth}
        \centering
        \includegraphics[width=\textwidth]{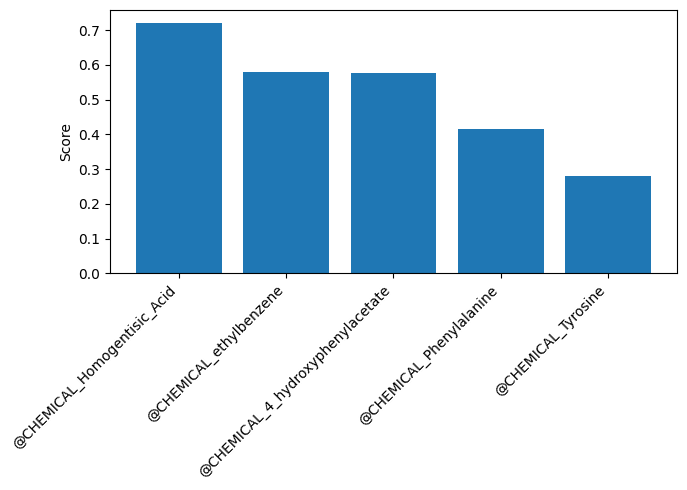}
        \caption{Five chemicals with non-zero HeteSim scores in the \textit{high-confidence network}.}
        \label{fig:hetesimhi}
    \end{subfigure}
    \begin{subfigure}[t]{0.48\textwidth}
        \centering
        \includegraphics[width=\textwidth]{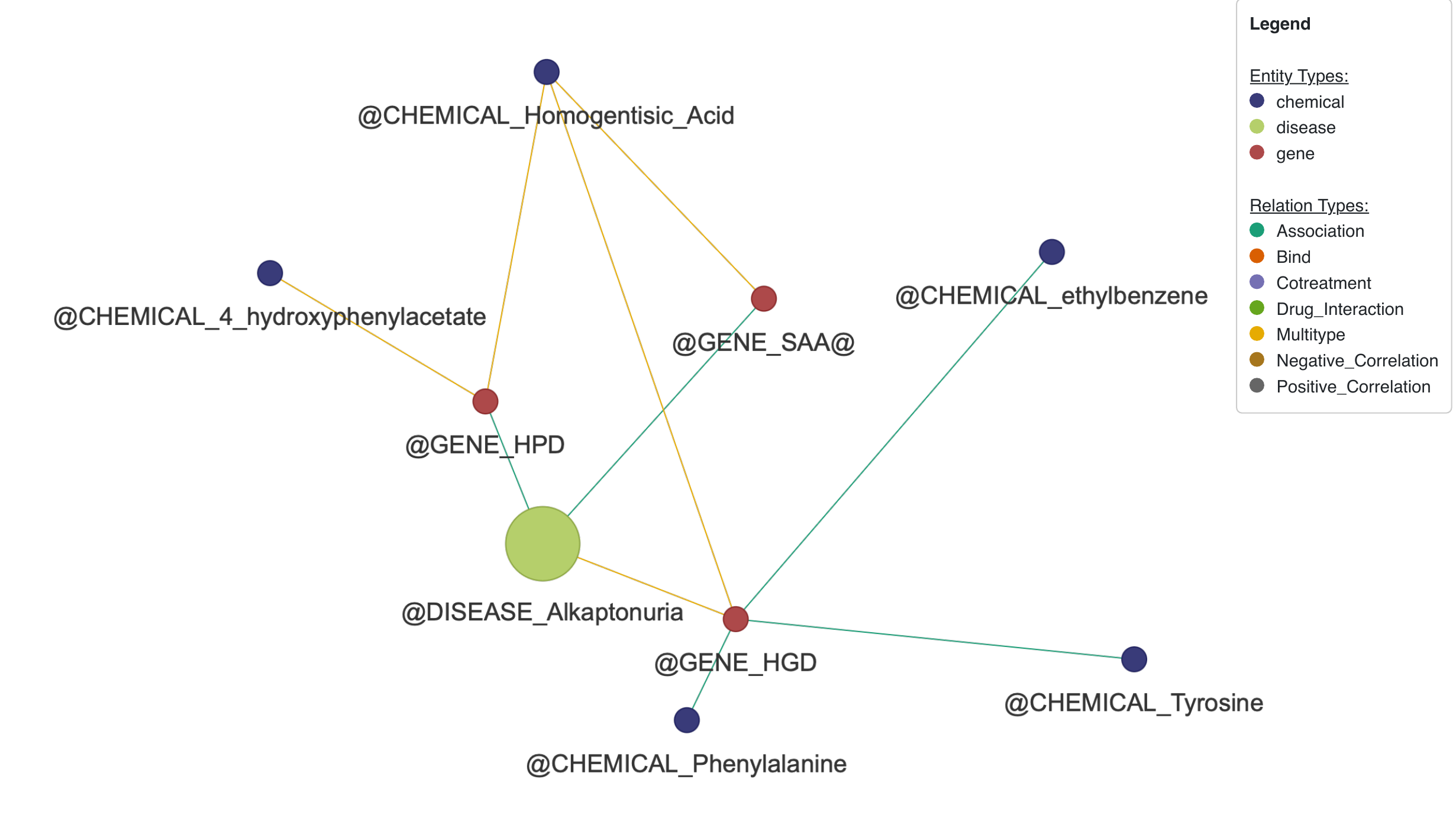}
        \caption{Meta-paths of the top 5 chemicals with non-zero HeteSim scores in the \textit{high-confidence network}. An HTML view of the meta-paths is available at \href{https://giangpth.github.io/Alkaptonuria/visualizations/hiHeteSimsubgraph.html}{link}.}
        \label{fig:hetesimhigraph}
    \end{subfigure}
    \caption{Top chemicals with the highest HeteSim scores with respect to \textit{Alkaptonuria}.}
    \label{fig:hetesim}
\end{figure}

To further explore chemicals potentially relevant to the target disease AKU, we performed a meta-path-based analysis, considering meta-paths of the form Chemical → Gene → Disease, and ranking chemicals based on the link to AKU (HeteSim score, see Methods for details). The top 20 chemicals are presented in Figure~\ref{fig:hetesim} for both networks, including also the meta-paths generated by them. For the high-confidence network, the set of entities with a non-zero HeteSim score was reduced; we include them all in the plot. In the \textit{extended network} (Figures \ref{fig:hetesimex} and \ref{fig:hetesimexgraph}), HGA emerges with the highest HeteSim score. This result consistently validates the approach, as HGA is the primary metabolite that accumulates in AKU. Other highly ranked chemicals include aromatic amino acids and related metabolites (e.g., phenylalanine-, tyrosine- and benzenoid-derived compounds), which are coherently embedded in the phenylalanine/tyrosine degradation pathway that is disrupted by the disease. The corresponding meta-paths demonstrate that these associations are predominantly mediated by HGD and related genes, suggesting that HeteSim accurately identifies biochemically relevant Chemical → Gene → Disease relationships. We observe again the presence of entities involved in bone remodeling, including genes TNFSF11,
SOST, DKK1, and a therapeutic intervention through Romosozumab. The inflammatory module is also included, confirming observations above.  In the \textit{high-confidence network} (Figures \ref{fig:hetesimhi} and \ref{fig:hetesimhigraph}), the number of chemicals with non-zero HeteSim scores is markedly reduced, reflecting the stricter evidence threshold. Only five compounds remain, all of which are directly connected to AKU via well-established metabolic or enzymatic links. The simplified meta-paths highlight a compact, biologically interpretable core centred on HGA and upstream aromatic amino acids. This reinforces the idea that the high-confidence KG retains primarily canonical, experimentally supported mechanisms.

\subsubsection*{Chemical and disease similarity}
\begin{figure}[ht]
    \centering
    \begin{subfigure}[t]{0.48\textwidth}
        \centering
        \includegraphics[width=\textwidth]{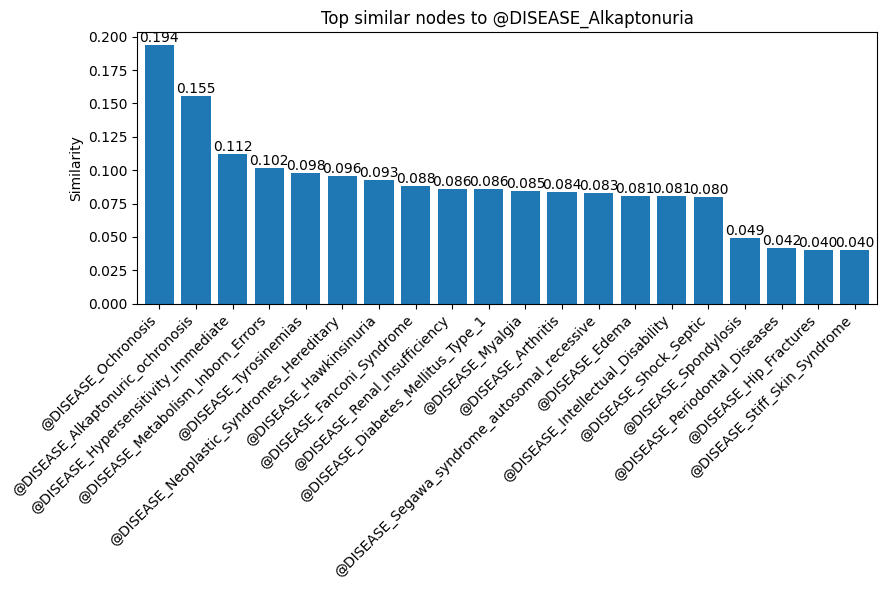}
        \caption{Top 20 diseases with the highest similarity scores  with respect to AKU  in the \textit{extended network}.}
        \label{fig:disimex}
    \end{subfigure}
    \hfill
    \begin{subfigure}[t]{0.48\textwidth}
        \centering
        \includegraphics[width=\textwidth]{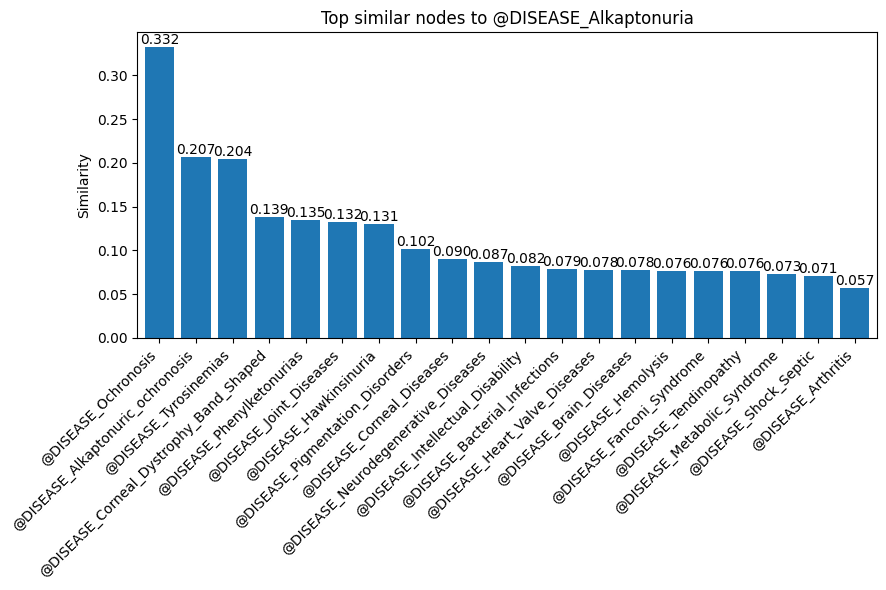}
        \caption{Top 20 diseases with the highest similarity scores  with respect to AKU  in the \textit{high-confidence network}.}
        \label{fig:disimhi}
    \end{subfigure}

    \centering
    \begin{subfigure}[t]{0.48\textwidth}
        \centering
        \includegraphics[width=\textwidth]{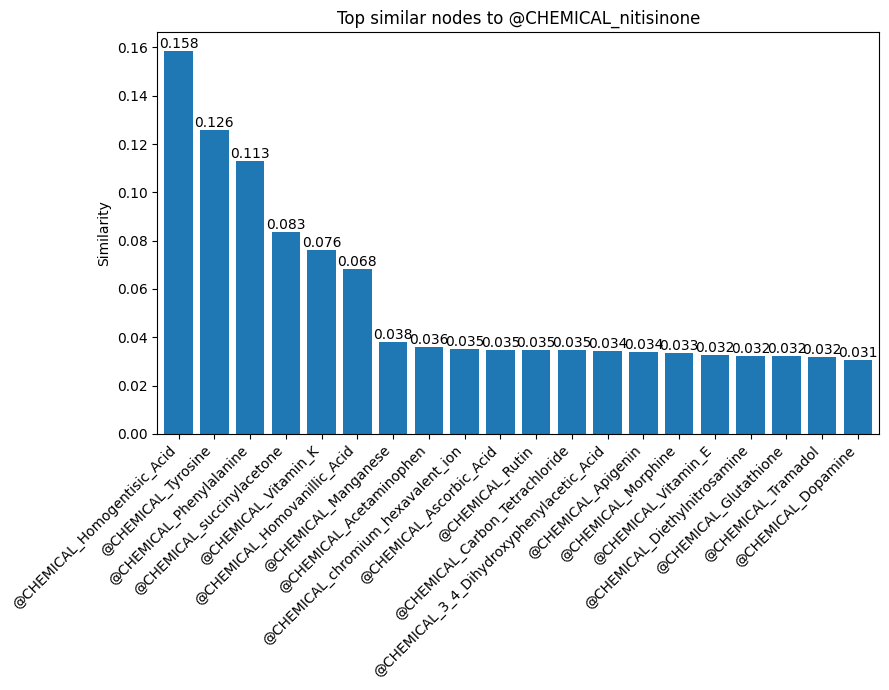}
        \caption{Top 20 chemicals with the highest similarity scores  with respect to nitisinone in the \textit{extended network}.}
        \label{fig:chemsimex}
    \end{subfigure}
    \hfill
    \begin{subfigure}[t]{0.48\textwidth}
        \centering
        \includegraphics[width=\textwidth]{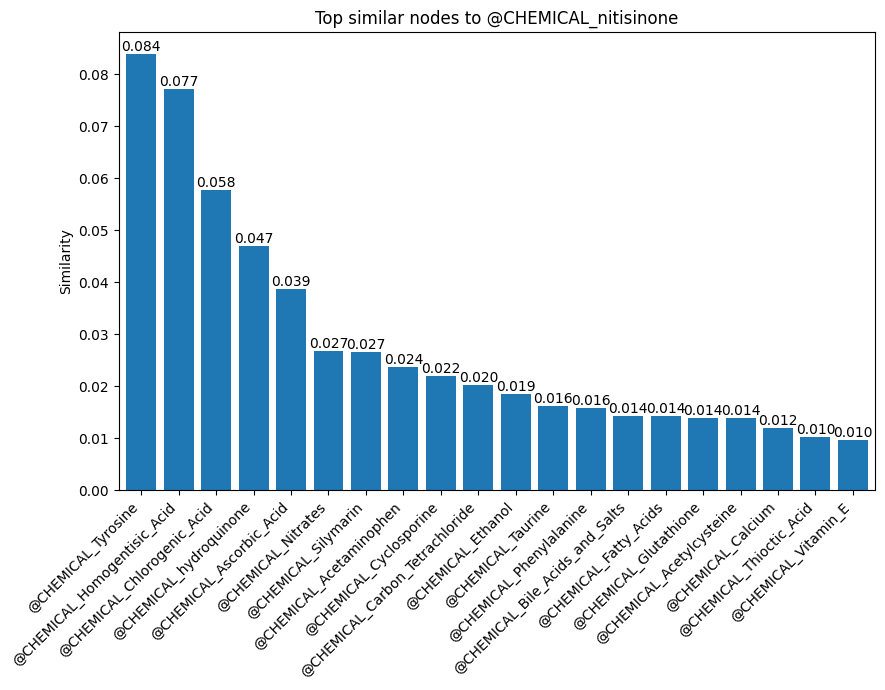}
        \caption{Top 20 chemicals with the highest similarity scores with respect to nitisinone in the \textit{high-confidence network}.}
        \label{fig:chemsimhi}
    \end{subfigure}
    
    \caption{Top diseases and chemicals similar to Alkaptonuria and nitisinone.}
    \label{fig:sim}
\end{figure}

A final analysis computed a similarity measure between diseases and chemicals, based on links to common entities. Figures~\ref{fig:disimex} and ~\ref{fig:disimhi} present the top diseases most similar to AKU, identified through the overlap of their connections with the same set of genes or chemicals. Given that \textit{nitisinone} is currently used in the treatment of AKU, we also performed a chemical similarity analysis using \textit{nitisinone} as the reference compound. Chemical similarity was determined based on shared connections to common genes or diseases. The top chemicals most similar to \textit{nitisinone} are shown in Figures~\ref{fig:chemsimex} and \ref{fig:chemsimhi}. The diseases most related to AKU (Figures~\ref{fig:disimex}  and \ref{fig:disimhi}) are similar to those identified using the centrality and meta-paths methods. These include chronic degenerative and inflammatory diseases. Therefore, with regard to AKU, rather than identifying new disease associations, these results demonstrate that the highest-ranking compounds represent the same molecular environment of chronic tissue damage and systemic remodeling, regardless of the chosen metric. Chemical similarity analysis using nitisinone as a reference compound (Figures~\ref{fig:chemsimex} and \ref{fig:chemsimhi}) also reproduced the previous results, with the highest-ranking compounds occupying related functional environments. Furthermore, the observation that there are relatively few similar chemicals in this \textit{high-confidence network} suggests that these connections are robust and specific.

\subsubsection*{Analysis of additional gene-gene and gene-drug interactions}

\begin{figure}[h!]
    \centering
    \includegraphics[width=0.8\linewidth]{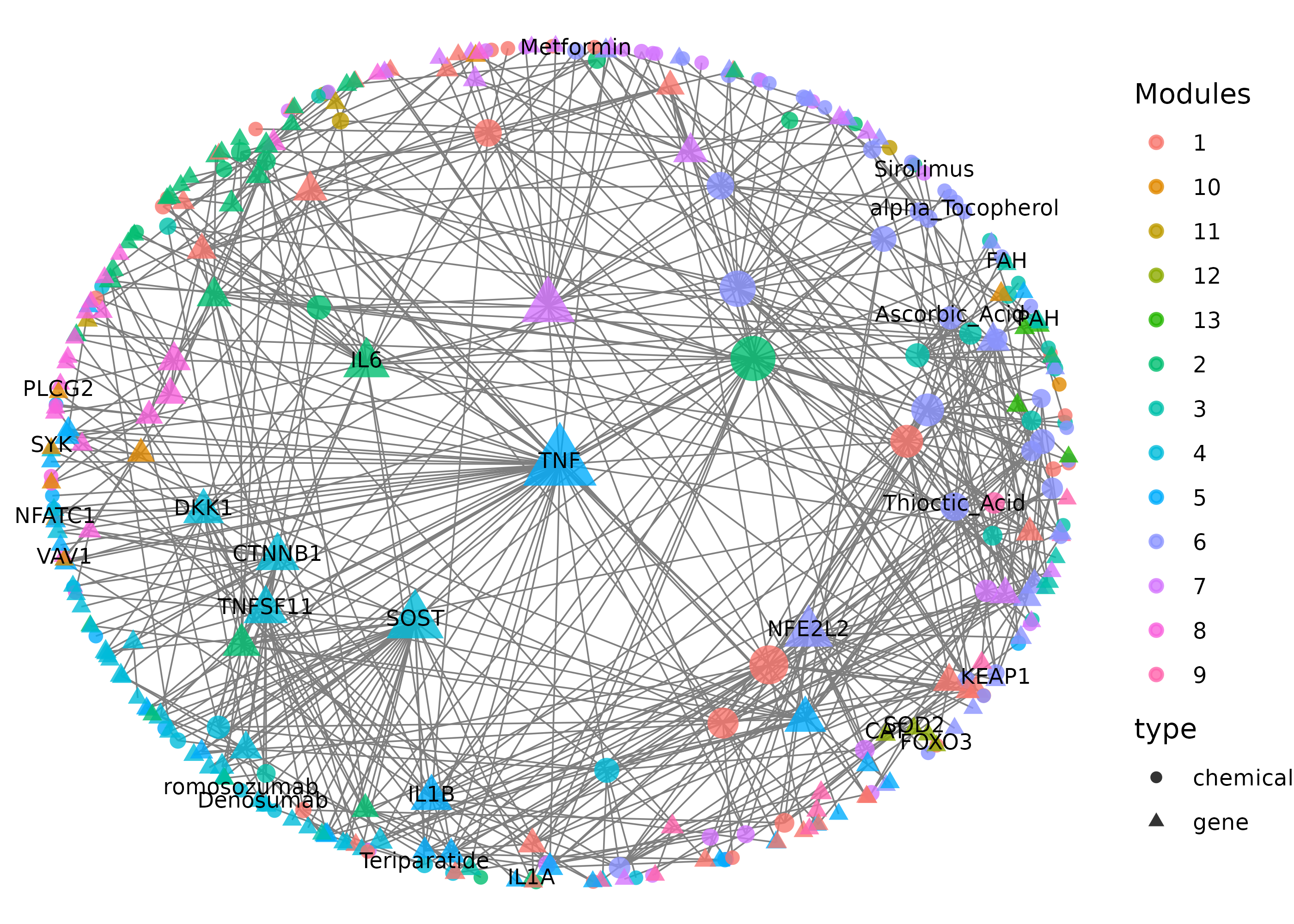}
    \caption{Main connected component of the network derived from combining the red and green edges of Figures~\ref{fig:histring} and \ref{fig:hidgidbgraph} (only genes and chemicals). }
    \label{fig:kg_giant_component_modules}
\end{figure}

The last step of our analysis concentrates on the red and green edges from Figures~\ref{fig:histring} and \ref{fig:hidgidbgraph}, combined into one single network. As noted before, green edges, showing interactions that are already present in the two databases, are a minority, therefore this analysis underlines the new knowledge present in our KG. We only consider genes and chemicals in order to perform functional enrichment analysis.  

The main connected component of this new network contains 349 nodes (206 genes and 143 chemicals) and can be divided into 13 modules (Figure \ref{fig:kg_giant_component_modules}). We include the complete enrichment analysis results as supplementary material on GitHub (see Data availability section). 

Functional enrichment analysis identified a distinct tyrosine/phenylalanine catabolism module (number 3) enriched for tyrosine metabolic processes (including FAH/PAH nodes), consistent with AKU’s causal defect in homogentisate metabolism. 

Two orthogonal modules converged on skeletal pathology: a bone development/WNT remodeling module (number 4) containing SOST, DKK1, CTNNB1, and TNFSF11, and an osteoclast signaling module (number 10) enriched for Fc receptor and SYK/PLCG2/VAV1/NFATC1 signaling pathways. These results align with reported osteoclastogenic imbalance and bone remodeling marker alterations observed in AKU cohorts~\cite{BRUNETTI20181059}. Interestingly in module 4 we can also observe the presence of Denosumab (a TNFRSF11 ligand inhibitor used for the management of osteoporosis in patients at high risk for bone fractures), Blosozumab (used in trials studying the basic science and treatment of Osteoporosis and Osteoporosis, Postmenopausal) and Romosozumab (a monoclonal antibody used to treat osteoporosis in postmenopausal women at high risk of fracture, patients who are intolerant of other treatments, or patients who have failed other treatments). These are “novel-to-AKU” in the sense of missing AKU-specific controlled trials, but are mechanistically proximal to the AKU osteoclastogenesis and bone remodeling findings (RANKL/OPG, sclerostin/DKK1). 
Always in module 4, there is also Teriparatide, a recombinant parathyroid hormone used for the treatment of osteoporosis, which can also be useful as a possible repurposable drug.

In parallel, inflammation and oxidative stress emerged as a recurrent theme, with immune/inflammatory modules (numbers 5, 6, 12) capturing TNF/IL1B-associated programs and oxidative stress modules enriched for NFE2L2/KEAP1 signaling and mitochondrial ROS defense genes (SOD2/CAT/FOXO3). This network structure reflects biomarker studies showing persistent elevation of inflammatory and oxidative stress markers in AKU patients (including SAA, IL6, IL1$\beta$, TNF$\alpha$, CRP), even in nitisinone-treated cohorts~\cite{BRACONI20181078}. 
In particular, module 5 with genes as NFKB1, RELA, NLRP3, CASP1, MAPK14 and MAPK1 enriches heavily for inflammatory response regulation and module 8 enriches for cytokine-mediated signaling and immune activation.

Module 6 is dominated by NRF2/KEAP1 detox programs (e.g., “Nuclear events mediated by NFE2L2”) and oxidative stress GO terms; module 12 highlights FOXO-mediated transcription and ROS metabolism. This aligns with AKU pigment chemistry evidence demonstrating radical-mediated collagen damage and mechanistic ties to redox imbalance~\cite{chow2020pigmentation}. Regarding these three modules, a plausible repurposable drug could be Dimethyl fumarate (DrugBank DB08908), a clinically used NRF2/KEAP1-pathway activator currently used in the relapsing-remitting form of multiple sclerosis. The same module also contains Ascorbate and Alpha Tocopherol. The former has already been tested in AKU human cells~\cite{spreafico2013antioxidants}, while the latter is a key antioxidant that protects cell membranes from oxidative damage.

Module 7’s top pathway terms are cancer signal programs since it contains PI3K/AKT1/EGFR/mTOR nodes; importantly, the enrichment analysis contains Autophagy(animal) and negative regulation of macroautophagy terms. This aligns with an AKU chondrocyte model demonstrating that HGA induces time-dependent autophagy alterations and culminates in chondroptosis with pigment accumulation~\cite{galderisi2022homogentisic}. Interestingly, linked to AKT1 and PRKAA2 (module 6), there is Metformin, commonly used for glycemic control in type 2 diabetes mellitus, that can inhibit the abnormal activation of the PI3K/AKT/mTOR signaling pathway in cartilage tissue, promote the restoration of cartilage cell autophagic function~\cite{xu2024metformin}. 
In this module, we also observe sirolimus, an mTOR inhibitor and immunosuppressant used to prevent organ transplant rejection.


\section*{Discussion}

In this work, we introduced a methodology for mining the scientific literature regarding a specific disease, using PubMed with the PubTator3 tool.
We aimed to derive crucial information about the genes, chemicals, drugs, and metabolites that are correlated to the disease, organized in an interaction network, i.e., a knowledge graph (KG). Our methodology also includes validation and knowledge extraction from the KG obtained.

We illustrated our work by studying the case of AKU. For rare diseases, usually there is less clinical data and research becomes more difficult and less supported.
Therefore, it becomes essential to have automatic tools to exploit the
knowledge in the scientific literature. 
We are effectively studying AKU and hence we already have some expertise as well as clinical data, which helped in the evaluation of our methodology.

We derived two KGs for AKU:  a very large network (called {\em extended network}) of correlations of genes, chemicals, and other diseases and a second, refined, smaller network, called {\em high-confidence network}, by considering only connections that are supported by at least two distinct research publications. This network includes less information (number of nodes reduced by a factor of 20) and can be more easily interpreted with respect to AKU. 

We validated our networks by comparing them to graphs extracted from the main public databases:  STRING  (for gene-gene interactions), DGIdb (for drug-gene interactions) and KEGG (for pathways). We observed that our KGs include biological knowledge on AKU similar to the three databases. Highest similarity was observed with KEGG, indicating that our KGs have a structure similar to biological pathways, where interactions include metabolites and other compounds.  Gene-gene and drug-gene relations, instead are rarely direct, but mostly captured by longer paths.

Besides including the existing knowledge on AKU, our KGs are more extensive, adding also new interactions and entities. We employed graph theory to study the two KGs and extract new hypotheses for important genes, pathways and possible therapies. In terms of general structure, we observed that the extended network is more clustered and less modular, while pruning out edges supported by only one publication allowed for a more modular network to emerge (the high-confidence network). This means that the high-confidence network may contain a structure that is more suitable for studying the disease, while the extended network could provide additional hypotheses. Both networks show a scale-free  structure with low degree assortativity typical to biological networks, where hub nodes connect many other less connected entities. 

In terms of communities, we have studied both networks, concentrating on the communities including AKU.    
We also identified the main genes, diseases and chemicals linked to AKU, from our KGs. We ranked entities based on different criteria, and inspected the top emerging ones. In all cases, known main associations were included, however also new genes, chemicals and diseases emerged.  In general, while the high-confidence network appeared to underline known metabolic effects of HGD mutations, the extended network provided connections to additional systemic effects and comorbidities. Among these, important connections were to oxidative stress and inflammatory pathways (with genes such as TTR and TNF in central positions), as well as bone remodeling (genes TNFSF11, SOST and DKK1, and therapies such as Romosozumab). Some indications of possible links between nitisinone and neurological disorders were also obtained.  

To further contextualise the new entities, through a more quantitative approach, we studied the additional interactions emerging from the comparison with  STRING  and DGIdb in one combined network. Here we performed a more systematic modularity and functional enrichment analysis, identifying again modules compatible to existing knowledge, but also new hypotheses. Specifically, we find again a strong connection to oxidative stress, inflammation and bone remodeling pathways, which is also observed in the analysis of the entire network. Furthermore, we found connections to chondrocyte autophagy. Drug repurposing hypotheses emerged from the analysis, including antioxidant and anti-inflammatory treatments such as Vitamin E (Alpha Tocopherol), Vitamin C, Thioctic Acid, bone-remodeling-related drugs (Romosozumab, Denosumab) and Metformin which can impact key regulators of chondrocyte metabolism. 

All in all, our work contributes (1) a general methodology for KG creation and validation  through literature mining,  (2) two different KGs on AKU, which include both existing knowledge on the disease and also new interactions that can be studied to generate new hypotheses and (3) a careful evaluation of new knowledge on AKU that can be extracted from the KGs. Future work will continue exploration of these networks, both computationally and in the laboratory, in particular the new pathways, genes and chemicals identified as possibly linked to AKU or useful in AKU management. 

\section*{Methods}
\subsection*{Data retrieval}
\label{sec:data_retrieval}
We employed PubMed to retrieve biomedical literature relevant to Alkaptonuria and PubTator3 \cite{wei2024pubtator} to find biomedical entities and relations among them from the retrieved literature. PubTator3 is a web-based text-mining resource developed and maintained by the National Center for Biotechnology Information (NCBI) to facilitate large-scale information retrieval from biomedical literature. It provides automatic annotation of PubMed and PubMed Central articles with key biological concepts such as genes, diseases, chemicals, variants, species, and cell lines. These annotations are generated using state-of-the-art named entity recognition and normalization methods, and are linked to standardized identifiers from major biomedical databases (e.g., MeSH, Entrez Gene, ChEBI, dbSNP).

\emph{Alkaptonuria}, \emph{HGD}, and \emph{homogentisic acid} were selected as the initial query terms for PubMed data retrieval. This process yielded 3,812 articles, from which PubTator3 extracted 4,435 unique named entities spanning categories such as diseases, chemicals, genes, and variants.  4,931 unique binary relations were also identified, each with a confidence score between 0.0 and 1.0, indicating the reliability of the underlying machine learning models. 

From this dataset, we extracted a secondary set of terms, to extend the initial network.  We considered all entities directly associated with Alkaptonuria, HGD, or homogentisic acid with an average confidence score greater than 0.7, resulting in 190 distinct entities beyond the three initial seeds. The generic terms Death and Disease were excluded from this secondary seed set, resulting in a total of 188 new query terms for PubMed. 

We queried PubMed again with each of these new entities. For entities associated with fewer than 2,000 PubMed records (PMIDs), all available articles were retrieved. For entities linked to more than 2,000 PMIDs, we limited the retrieval to a maximum of 2,000 articles published within the last five years to ensure recent and manageable data coverage. To focus on primary research studies, we applied the following PubMed filter:
\begin{verbatim}
{query_term} AND (Journal Article[pt] OR Clinical Trial[pt] 
OR Case Reports[pt] OR Randomized Controlled Trial[pt]
OR Observational Study[pt] OR Comparative Study[pt] 
OR Evaluation Study[pt])
NOT (Review[pt] OR Systematic Review[pt] 
OR Meta-Analysis[pt] OR Editorial[pt] 
OR Letter[pt] OR Comment[pt])
\end{verbatim}
This filter excludes secondary literature such as reviews or commentaries, retaining only original research articles.
In total, 168,502 PMIDs were retrieved, of which 165,785 were unique compared to the initial set of 3,812 PMIDs obtained from the three starting seeds (Alkaptonuria, HGD, and homogentisic acid).

PubTator3, applied to the new publications, yielded 68,702 unique entities and 360,026 unique relations. From this corpus, we applied a series of filtering steps. Relations categorized as Comparison were discarded. We retained only those relations with either (i) an average confidence score greater than 0.7, or (ii) an average confidence score greater than 0.5 supported by more than three publications. Furthermore, variants were systematically associated with their corresponding genes. 
The disease and chemical list returned by PubTator3 contained a number of overly general or non-informative entries (e.g., Death, Disease, Fat or Water), often highly connected in the network. To reduce the noise brought by these nodes, we manually filtered, based on domain expertise, those with high degree centrality (larger than 10), along with their associated relations. Entities that were not linked with the connected component containing Alkaptonuria were discarded. The resulting data were used to construct a connected undirected network, referred to as the \textit{extended network}, comprising 27,252 nodes (4,754 diseases, 10,792 chemicals, 11,534 genes and 172 variants associated with no gene) and 261,649 edges. Each edge was assigned a weight proportional to the number of supporting publications for the corresponding connection. Specifically, the weight $w_e$ of an edge $e$ was defined as $w_e = 1 -2^{-n}$, where $n$ denotes the number of distinct articles supporting the connection. This weighing choice allows to consider more strongly the connections supported by a large amount of literature, with a fast growth of the score towards the maximum of 1  as the number of supporting papers increases.   

From the extended network, we further derived a \textit{high-confidence network}, retaining only connections supported by at least two independent publications and considering only the connected component containing Alkaptonuria. The resulting network was substantially smaller, comprising 1,450 nodes (366 diseases, 602 chemicals and 482 genes) and 3,307 edges. 

The two networks were first compared with existing biomedical knowledge from public databases, for validation purposes. Subsequently, we employed network theory tools to characterize the network and extract additional knowledge on AKU.

\subsection*{Validation}
To validate our networks, we compare them against reference networks constructed from well-established, curated databases. These curated resources provide high-quality interaction data that serve as reliable benchmarks for evaluating the accuracy and biological relevance of our inferred networks. 

\subsubsection*{ STRING }
We begin the validation using  STRING \footnote{https://string-db.org}(https://string-db.org), a comprehensive database and web resource providing experimentally supported and curated gene–gene interactions. The validation procedure is as follows: gene entities were extracted from our networks and queried in  STRING . Since  STRING  integrates connections from various sources, including predictive models and text mining, we restricted the results to interactions supported exclusively by \textit{experimental evidence} and \textit{curated databases}. The retrieved interactions were then compared against those in our network. We also compared to a second set of subnetworks from our networks, obtained by computing the shortest paths between genes directly connected in  STRING  and not in our networks. This is due to the fact that we observed that genes in our network are rarely directly connected, but they may have diseases or chemicals in common. 

Given the large size of the \textit{extended network}, validation was conducted only on the module containing \textit{Alkaptonuria}, \textit{HGD} and \textit{Homogentisic acid}. This module consists of 4,008 nodes, including 974 genes. 

\subsubsection*{DGIdb: Drug-gene interactions}
Following the same procedure as with  STRING , we extracted gene entities from our networks to formulate queries for the \textit{Drug–Gene Interaction Database} (DGIdb)~\cite{cannon2024dgidb}(https://dgidb.org). The resulting gene–drug interactions were then compared against those identified in our networks.

\subsubsection*{Tyrosine pathway from KEGG}
To improve consistency with established biomedical knowledge on \textit{AKU}, we compared the \textit{extended network} with the human tyrosine metabolism pathway curated in \textit{KEGG}\cite{kanehisa2000kegg}(https://www.kegg.jp/entry/hsa00350). From this canonical pathway, we constructed a reference metabolic network by defining edges according to three types of biochemical associations: (i) connections between compounds that act as co-reactants in the same enzymatic reaction, (ii) links from reactant(s) to their corresponding product(s), and (iii) associations between substrates and their catalyzing enzymes, when available. The curated metabolic network constructed from KEGG pathway serves as a reference framework for evaluating the biological relevance of the KGs extracted by our methodology. 

\subsection*{Knowledge extraction}
After validating and comparing our KGs with existing knowledge, we use network theory tools to study the two networks.

\paragraph{Network Characterization}
We start by computing general metrics on the two networks \cite{10.1093/oso/9780198805090.001.0001}, to study the general structure of the two KGs, as follows. The network diameter gives an indication of how globally connected the nodes are by paths, on average. The degree distribution also provides indications on the structure of the network. The most central nodes, in the sense that they are closest, on average in the worst case, to all other nodes, are extracted. Clustering coefficients investigate local connectivity. Assortativity is investigated using degree correlation.

\paragraph{Community detection}

In large-scale biological networks, community detection plays a crucial role in uncovering the modular organization of entities that underlie complex cellular processes. Genes or proteins within the same community are often functionally related or participate in similar biological mechanisms, making modularity-based methods particularly suitable for systems biology and network medicine. Among these, modularity optimization is one of the most widely used approaches, as it quantifies how well a given network partition separates dense intra-community connections from sparse inter-community ones \cite{newman2004finding}. In this study, we use the Leiden algorithm \cite{traag2019louvain} (with resolution parameter of 1.0 for standard modularity detection) to discover the modularity-based communities in our networks.

Besides modularity-based methods, alternative ways to explore the community around a certain node are looking at k-cores and cliques 
\cite{10.1093/oso/9780198805090.001.0001}. A \textit{k-core} is a maximal subgraph in which every node has a degree of at least \textit{k} within that subgraph. The \textit{maximum k-core} associated with a given node represents the highest value of \textit{k} for which the node remains part of a k-core. This structure captures the most cohesive neighborhood surrounding the node by filtering out loosely connected elements and retaining only its most stable and mutually reinforced connections.  Cliques, on the other hand, are groups of nodes that are completely connected with each other. They reveal highly connected structures within a network. In this work we compute the maximum k-core subgraph and the maximum clique for the node corresponding to AKU, in order to detect entities closely connected to the disease in our network.

\paragraph{Entities related to Alkaptonuria}
We explore advanced centrality measures that could identify important entities in the disease under study. In particular, we employ Personalized PageRank \cite{pagerank1998} and Personalized Katz \cite{Katz_1953} centrality to rank entities based on their importance in the network, having Alkaptonuria as a central start node.  At its core, PageRank imagines a ``random surfer'' who moves through the network by following links. Over time, the frequency with which the surfer lands on a node reflects that node’s overall importance in the network. Personalized PageRank adapts this idea by introducing a bias: instead of starting from anywhere in the network, the random surfer has a higher probability of returning to a specific set of nodes that represent the user’s interest. This makes the measure ``personalized'' since the importance of each node is evaluated relative to the chosen starting points, not in the network as a whole. We set Alkaptonuria as the starting point in the random walk. 

While PPR emphasizes nodes within the immediate functional neighborhood of \textit{Alkaptonuria}, we also applied \textit{Personalized Katz centrality} \cite{Katz_1953} to capture the diffusion of influence across more distant pathways, using a large decay factor of $0.95 / \lambda_{max}$. This approach integrates both local and global relational information within the network. Furthermore, we employ meta-path analysis, considering meta-paths of the form Chemical → Gene → Disease. For each chemical, we then computed the HeteSim score (Heterogeneous Similarity) \cite{shi2014hetesim} with respect to Alkaptonuria.

\paragraph{Chemical and disease ranking by genetic similarity to AKU}
{The similarity between diseases is computed based on the genes and/or chemicals to which they are connected in the network. Diseases are considered similar if they share a substantial number of associated genes or chemicals, with greater emphasis placed on associations that are specific rather than ubiquitous across diseases. Similarity scores derived from gene-based and chemical-based evidence are integrated, and weak associations are filtered by retaining only the strongest connections. This procedure yields a matrix in which entries quantify pairwise disease similarity. An analogous approach is applied to compute similarity between chemicals.
}

\paragraph{Analysis of additional gene-gene and gene-drug interactions}
We conclude the knowledge-extraction phase of our study with a more in-depth analysis of interactions included in our KGs and not in existing databases, emerging from the validation analysis presented earlier. The objective is to provide a more quantitative evaluation of the new biological knowledge that can be extracted.  Specifically, we concentrate on nodes present in  STRING  and DGIdb, i.e., the subgraphs used in the validation analysis, and mainly consider the edges that are present in our high-confidence network and that are not present in the two databases. To connect to existing knowledge, we also include interactions present both in our high-confidence network and in the two databases.  Analyses were performed on the giant connected component of the merged network. We employ only genes and chemicals to enable functional enrichment analysis. Community structure was identified via modularity optimization (fast-greedy clustering)\cite{igraph2023}. Then, for each module, gProfiler~\cite{gProfiler} was used to evaluate the genes and identify enriched pathways and processes. We used adjusted p-values through standard \emph{g:SCS} correction.

\bibliography{sample}



\section*{Acknowledgements}
This work was supported by the co-funding European Union - Next Generation EU, in the context of The National Recovery and Resilience Plan, Mission 4 Component 2, Investment 1.1, Call PRIN 2022 D.D. 104 02-02-2022 – MEDICA Project, CUP N. I53D23003720006 \& B53D23013170006. We also acknowledge partial support from Investment 1.5 Ecosystems of Innovation, Project Tuscany Health Ecosystem (THE), CUP: B83C22003920001, Spoke 3, from the SPARK programme at the University of Pisa, from the project "Hub multidisciplinare e interregionale di ricerca e sperimentazione clinica per il contrasto alle pandemie e all’antibioticoresistenza (PAN-HUB)” funded by the Italian Ministry of Health (POS 2014-2020, project ID: T4-AN-07, CUP: I53C22001300001), from a 2023 NARSAD Young Investigator Grant from the Brain \& Behavior Research Foundation, and from the Italian MUR PRIN PNRR 2022 project ``DELICE'' \emph{Decentralized Ledgers in Circular Economy} (P20223T2MF),.

\section*{Data availability}
Analysis code is available from \href{https://github.com/giangpth/Alkaptonuria.git}{GitHub}\footnote{https://github.com/giangpth/Alkaptonuria.git}. The two KGs are made available as supplementary data, in JSON format, also on \href{https://github.com/giangpth/Alkaptonuria/blob/main/data}{GitHub}\footnote{https://github.com/giangpth/Alkaptonuria/blob/main/data/extended\_graph.json}$^,$\footnote{https://github.com/giangpth/Alkaptonuria/blob/main/data/highconf\_graph.json}. Full functional enrichment analysis results are available from the same \href{https://github.com/giangpth/Alkaptonuria/blob/main/Enrichment.zip}{ repository}\footnote{https://github.com/giangpth/Alkaptonuria/blob/main/Enrichment.zip}. The figure captions contain links to additional visualizations for each figure, in HTML format. The PubMed data used in this work is available online on the NCBI website.  

\section*{Author contributions statement}

L.B., S.B., M.F., P.M., C.P., O.S. and A.S.(2) conceived the experiment; All authors designed and revised the analysis pipeline; G.P. extracted PubMed data, wrote the code, built the KGs, ran the graph theory analysis and generated the visualizations; S.G.G. performed the analysis of additional edges in the KGs through clustering and functional enrichment; R.F., B.R., A.B., S.G.G., A.S.(1), O.S. and A.S.(2) performed the biomedical analysis and interpretation of results; C.G. performed the related literature analysis. All authors interpreted the results and contributed to drafting the manuscript. All authors approved the final manuscript.

\section*{Additional information}

\subsection*{Competing interests} All authors declare no competing interests. 





\end{document}